\documentclass[journal]{IEEEtran}

\usepackage{amsmath}   
\usepackage{amssymb}   
\usepackage{bm}        
\usepackage{microtype} 
\usepackage{graphicx}  
\usepackage{tikz}      
\usepackage{booktabs}  
\usepackage{multirow}  
\usepackage{array}     
\usepackage{hyperref}  
\usepackage{cite}      
\usepackage{float}
\usepackage{enumitem}
\usepackage[table]{xcolor}
\definecolor{cyanblue}{RGB}{224,238,255}
\usepackage{graphicx}
\usepackage{subcaption}
\graphicspath{{figures/}} 
\usepackage{tcolorbox}
\usepackage[ruled,vlined]{algorithm2e}
\usepackage{booktabs}
\usepackage{siunitx}
\usepackage{longtable}
\newcommand*\halfcirc[1][1ex]{%
  \begin{tikzpicture}
  \draw[fill] (0,0)-- (90:#1) arc (90:270:#1) -- cycle ;
  \draw (0,0) circle (#1);
  \end{tikzpicture}}
\newcommand*\fullcirc[1][1ex]{\tikz\fill (0,0) circle (#1);}
\usepackage{tikz}
\usepackage[edges]{forest}
\usepackage{xcolor}
\usetikzlibrary{calc}

\usepackage{adjustbox}
\usepackage{caption}
\renewcommand{\arraystretch}{1.1} 
\usepackage{amssymb}    
\usepackage{pifont}     
\usepackage{xcolor}     
\newcommand*\emptycirc[1][1ex]{\tikz\draw (0,0) circle (#1);} 
\newcolumntype{Y}{S[table-format=1.3]}

\usepackage{bm}   
\usepackage{listings}
\usepackage{xcolor} 
\usepackage{listings}
\usepackage{xcolor}
\usepackage{xcolor}
\usepackage[T1]{fontenc}
\usepackage[utf8]{inputenc} 
\usepackage{listings}
\usepackage{xcolor}
\usepackage[T1]{fontenc}
\usepackage[utf8]{inputenc} 

\lstdefinelanguage{json}{
  morestring=[b]",
  moredelim=[s][\color{black}]{\{}{\}},
  moredelim=[s][\color{black}]{[}{]},
  stringstyle=\color{brown},
  showstringspaces=false,
}

\lstdefinestyle{jsonschema}{
  language=json,
  basicstyle=\ttfamily\footnotesize,
  frame=single,
  breaklines=true,
  columns=fullflexible,
  keepspaces=true,
  showstringspaces=false,
}
\lstdefinestyle{plain}{
  basicstyle=\ttfamily\footnotesize,
  frame=single,
  breaklines=true,
  columns=fullflexible,
  keepspaces=true,
  showstringspaces=false,
}

\lstdefinestyle{python}{
  language=Python,
  basicstyle=\ttfamily\footnotesize,
  frame=single,
  breaklines=true,
  columns=fullflexible,
  keepspaces=true,
  showstringspaces=false,
  keywordstyle=\color{blue},
  commentstyle=\color{gray},
  stringstyle=\color{brown},
}
\newcommand{\vect}[1]{\bm{#1}}   

\ifCLASSINFOpdf
\else
\fi
\begin{document}
%
\title{\textit{UAVBench}: An Open Benchmark Dataset for Autonomous and Agentic AI UAV Systems via LLM-Generated Flight Scenarios}

\author{
Mohamed~Amine~Ferrag$^{*\S}$,~\IEEEmembership{Senior~Member,~IEEE}, 
Abderrahmane~Lakas$^{*}$,~\IEEEmembership{Senior~Member,~IEEE},
and~Merouane~Debbah$^{1}$,~\IEEEmembership{Fellow,~IEEE}%
\thanks{$^{*}$Department of Computer and Network Engineering, College of Information Technology, United Arab Emirates University, Al Ain, United Arab Emirates.}%
\thanks{$^{1}$Khalifa University of Science and Technology, Abu Dhabi, United Arab Emirates.}%
\thanks{$^{\S}$Corresponding author: \texttt{mohamed.ferrag@uaeu.ac.ae}}%
}

%
%

\markboth{ }%
{Shell \MakeLowercase{\textit{et al.}}: Bare Demo of IEEEtran.cls for IEEE Journals}
%



\maketitle


\begin{abstract}
Autonomous aerial systems increasingly rely on large language models (LLMs) for mission planning, perception, and decision-making; yet, the lack of standardized, physically grounded benchmarks limits systematic evaluation of their reasoning capabilities. To address this gap, we introduce UAVBench, an open benchmark dataset comprising \textit{50,000 validated UAV flight scenarios} generated through taxonomy-guided LLM prompting and multi-stage safety validation. Each scenario is encoded in a structured JSON schema encompassing mission objectives, vehicle configuration, environmental conditions, and quantitative risk labels, providing a unified representation of UAV operations across diverse domains. Building on this foundation, we present UAVBench\_MCQ, a reasoning-oriented extension containing \textit{50{,}000 multiple-choice questions} spanning ten cognitive and ethical reasoning styles—from aerodynamics and navigation to multi-agent coordination and hybrid integrated reasoning. This framework enables interpretable, machine-checkable assessment of UAV-specific cognition under realistic operational contexts. We evaluate 32 state-of-the-art LLMs, including GPT-5, ChatGPT~4o, Gemini~2.5~Flash, DeepSeek~V3, Qwen3~235B, and ERNIE~4.5~300B, and find strong performance in perception and policy reasoning but persistent challenges in ethics-aware and resource-constrained decision-making. \texttt{UAVBench} establishes a reproducible, physically grounded foundation for benchmarking agentic AI in autonomous aerial systems and advancing next-generation UAV reasoning intelligence. To support open science and reproducibility, we release the \texttt{UAVBench} dataset (including labeled data), the \texttt{UAVBench\_MCQ} benchmark, evaluation scripts, and all related materials on GitHub: \url{https://github.com/maferrag/UAVBench}.
\end{abstract}


%
\IEEEpeerreviewmaketitle

\section{Introduction}

Large Language Models (LLMs) are emerging as powerful tools for enhancing UAV autonomy. Recent studies have increasingly explored integrating LLMs into UAV systems to improve autonomy, decision-making, and communication. Several works demonstrate how LLMs can augment or replace traditional reinforcement learning and optimization frameworks, which often struggle with training complexity and low sample efficiency \cite{ferrag2025reasoning}. For example, LLMs have been applied to the Internet of Drones via hybrid decision-making frameworks that combine discovery Generation with structured knowledge graphs, allowing interpretable context-aware UAV control \cite{sezgin2025llm}. Other efforts employ LLM-guided reinforcement learning to address security and energy-efficiency trade-offs in heterogeneous UAV networks, achieving improved secrecy rates and robust trajectory optimization \cite{zheng2025llm}. Similarly, LLM-based in-context learning has been introduced for intelligent data collection scheduling in UAV-assisted networks, outperforming baseline strategies while also revealing vulnerabilities to adversarial manipulation \cite{emami2025llm}. Additional work focuses on minimizing the age of information in UAV-assisted sensor networks using an evolutionary-optimization-assisted LLM, demonstrating superior routing efficiency under high node density conditions \cite{wei2025laura}. These developments highlight LLMs' ability to infuse adaptability, interpretability, and semantic reasoning into UAV decision pipelines.

Beyond single-agent autonomy, LLMs are also being leveraged for multi-agent UAV coordination and large-scale operational contexts. Recent frameworks utilize iterative structured prompting to optimize multi-hop UAV placements, thereby reducing computational overhead while maintaining near-optimal performance in network backhaul scenarios \cite{wang2025multi,li2025efficient}. Other works demonstrate the effectiveness of LLM-based in-context learning for flight resource allocation in wildfire monitoring, where real-time scheduling is critical to minimizing latency and data staleness \cite{emami2025frsicl}. Hierarchical architectures that combine high-altitude platforms and onboard UAV LLMs have been proposed for 3D aerial highway systems, providing both strategic access control and tactical maneuvering \cite{yan2025hierarchical}. In the domain of swarm intelligence, LLM-driven role-adaptive frameworks enhance collaboration through semantic communication and dynamic role switching, improving task coverage and generalization in multi-UAV systems \cite{wang2025rally}. Likewise, urban trajectory planning approaches merge DRL with LLM reasoning to ensure safe, efficient, and regulation-compliant operations in low-altitude economic airspaces \cite{gong2025safe}. Together, these works demonstrate that LLMs are not merely auxiliary tools but are emerging as core enablers of intelligent, interpretable, and scalable UAV autonomy across diverse mission profiles \cite{sekaran2025urbaning}.

Constructing intelligent agents capable of understanding natural language commands and translating them into navigation behaviors remains a central challenge in artificial intelligence \cite{zheng2025uav}. Although vision-language navigation (VLN) has been extensively studied for ground robots, the aerial domain introduces far greater complexity. UAVs must operate within continuous three-dimensional environments characterized by high degrees of freedom, varying altitudes, dynamic obstacles, and fluid environmental conditions such as wind and lighting changes \cite{yuan2025next}. These factors make path planning, spatial reasoning, and language grounding considerably more difficult than in ground-based systems. Moreover, aerial navigation demands fine-grained control over orientation, velocity, and stability, where small errors can lead to mission failure or collisions. Consequently, direct adaptations of ground-based VLN methods—typically optimized for discrete, planar movements—fail to capture the continuous, physics-driven nature of aerial motion and the real-time decision-making constraints inherent to flight \cite{wang2024towards}.

Despite growing research attention, existing UAV VLN benchmarks and datasets remain limited in terms of realism, task diversity, and physical grounding. Many rely on simplified discrete actions, static environments, or low-fidelity simulators that neglect the continuous control challenges central to UAV operation \cite{li2025efficient}. This lack of physical and semantic richness hinders progress toward fully embodied aerial intelligence. Bridging this gap requires specialized platforms and datasets that integrate realistic flight dynamics, multimodal perception, and mission-level reasoning. Such resources would not only enable more accurate simulation of UAV flight behaviors but also facilitate the study of complex reasoning and language-grounded decision-making under real-world constraints. Addressing these limitations motivates the development of unified benchmarks such as UAVBench, which couple scenario-level realism with structured reasoning evaluation, paving the way for end-to-end research on autonomous, language-guided aerial navigation.

Unmanned aerial vehicles (UAVs) are increasingly deployed across domains, including disaster response, agriculture, environmental monitoring, traffic observation, and energy infrastructure inspection. However, most missions still depend on human-operated remote control, which is labor-intensive, error-prone, and costly \cite{yao2024aeroverse}. Developing autonomous UAV agents that can perceive, reason, and act in complex environments is therefore a critical research objective. Compared to ground-based or indoor agents, UAVs face distinct challenges such as operating in large-scale, dynamic 3D environments, managing costly data collection, and requiring well-defined aerial-embodied tasks. Addressing these challenges requires specialized simulators, datasets, and evaluation frameworks that facilitate training and benchmarking of UAV embodied intelligence \cite{guo2025bedi}.

Beyond task execution, UAV autonomy requires advancing from fine-grained instruction-based navigation to high-level, goal-oriented cognition. Emerging approaches, such as Object Goal Navigation (ObjectNav), demonstrate the potential of semantic-driven navigation, in which agents reach mission-critical targets using abstract goals rather than detailed step-by-step instructions \cite{xiao2025uav}. Although ObjectNav has been explored in indoor ground settings, its application to outdoor aerial environments remains underdeveloped. At the same time, human-like embodied cognition —processing continuous first-person visual streams for orientation, reasoning, and navigation —is largely absent from current UAV research. Urban airspaces, with their vertical mobility, dynamic obstacles, and dense semantic complexity, present new challenges for autonomous navigation. To advance the field, it is imperative to establish systematic, standardized, and open benchmarks that evaluate the cognition embodied in UAVs and enable robust, scalable autonomy in real-world scenarios \cite{zhao2025urbanvideo}.

Our study is guided by the following research questions, designed to investigate how structured, physically grounded UAV scenarios and reasoning-based evaluation frameworks can advance the development of autonomous aerial intelligence:

\begin{tcolorbox}[
    colback=gray!10,
    colframe=black,
    arc=6pt,
    boxrule=0.7pt,
    left=2mm, right=2mm, top=1mm, bottom=1mm,
    title=Research Questions
]
\small
\begin{itemize}
    \item \textbf{RQ1:} How can a unified schema and taxonomy-driven generation framework ensure that large-scale UAV scenarios remain physically consistent, safety-aware, and semantically diverse for benchmarking autonomous flight intelligence?

    \item \textbf{RQ2:} What methods can be employed to systematically validate and risk-label automatically generated UAV scenarios to guarantee physical feasibility, schema compliance, and interpretability?

    \item \textbf{RQ3:} How can structured reasoning tasks derived from validated UAV scenarios be formulated to evaluate and compare cognitive, ethical, and operational decision-making in autonomous aerial systems?

    \item \textbf{RQ4:} How do distinct reasoning styles—spanning physical, navigational, ethical, and hybrid domains—affect the accuracy, generalization, and reliability of intelligent agents when performing UAV-related reasoning tasks?

    \item \textbf{RQ5:} To what extent do different model architectures and training paradigms influence consistency and grounded reasoning performance across diverse UAV mission contexts?
\end{itemize}

\end{tcolorbox}

To address these research questions, we introduce \texttt{UAVBench}, an open benchmark dataset constructed from LLM-generated UAV flight scenarios for evaluating and training agentic AI models in autonomous aerial systems. \texttt{UAVBench} unifies scenario generation, validation, risk labeling, and reasoning into a single framework that systematically produces structured and physically consistent UAV missions. Each scenario is represented as a validated JSON specification capturing the UAV configuration, environment, mission objectives, airspace geometry, and safety constraints. The dataset integrates a multi-stage validation pipeline to ensure schema compliance, physical feasibility, and hazard-aware labeling, thereby enabling large-scale benchmarking of autonomous flight intelligence. Furthermore, we extend this dataset with \texttt{UAVBench-MCQ}, a reasoning-oriented benchmark that evaluates the cognitive, ethical, and operational decision-making capabilities of large language models (LLMs) in UAV contexts. The key contributions of this work are summarized as follows:

\begin{table*}[ht]
\centering
\caption{Comparative coverage analysis of UAV embodied-intelligence benchmarks.}
\label{tab:uavbench_comparison_matrix}
\scriptsize
\setlength{\tabcolsep}{5pt}
\renewcommand{\arraystretch}{1.05}

\newcolumntype{C}{>{\centering\arraybackslash}m{0.08\linewidth}}
\rowcolors{2}{white}{cyanblue!70} 
\begin{tabular}{p{4cm} c C C C C C C C}  
\toprule
\rowcolor{gray!25}
\textbf{Work} & \textbf{Year} &
\multicolumn{3}{c}{\textbf{Scenario Design}} &
\multicolumn{3}{c}{\textbf{Reasoning Scope}} &
\textbf{Evaluation} \\
\cmidrule(lr){3-5} \cmidrule(lr){6-8}
 &  & Physical Realism & Validation \& Risk & Mission Diversity &
 Physics/Navigation & Ethics/Safety & Hybrid Reasoning &
 Structured MCQs \\
\midrule
Wang et al.~\cite{wang2024towards} (OpenUAV) & 2024 & \fullcirc & \halfcirc & \emptycirc & \fullcirc & \emptycirc & \emptycirc & \emptycirc \\
Yao et al.~\cite{yao2024aeroverse} (AeroVerse) & 2024 & \fullcirc & \halfcirc & \fullcirc & \fullcirc & \emptycirc & \emptycirc & \emptycirc \\
Guo et al.~\cite{guo2025bedi} (BEDI) & 2025 & \fullcirc & \emptycirc & \fullcirc & \halfcirc & \emptycirc & \emptycirc & \emptycirc \\
Xiao et al.~\cite{xiao2025uav} (UAV-ON) & 2025 & \fullcirc & \emptycirc & \halfcirc & \fullcirc & \emptycirc & \emptycirc & \emptycirc \\
Zhao et al.~\cite{zhao2025urbanvideo} (UrbanVideo-Bench) & 2025 & \halfcirc & \emptycirc & \emptycirc & \halfcirc & \emptycirc & \emptycirc & \halfcirc \\
\textbf{UAVBench / \texttt{UAVBench\_MCQ}} & 2025 & \fullcirc & \fullcirc & \fullcirc & \fullcirc & \fullcirc & \fullcirc & \fullcirc \\
\bottomrule
\end{tabular}\\
Symbols denote coverage levels:  \fullcirc = fully covered,  \halfcirc = partially covered,  \emptycirc = not covered.
\end{table*}

\begin{enumerate}
  \item \textit{Unified UAV Scenario Schema:} 
  We propose a structured and mathematically defined schema that represents each UAV mission as a tuple encompassing simulation dynamics, vehicle configuration, environmental conditions, mission planning, and safety constraints. This schema ensures consistency, physical validity, and interoperability across diverse UAV applications.

  \item \textit{Taxonomy-Guided Scenario Generation:} 
  We develop a taxonomy-driven LLM prompting mechanism that samples from a factorized space of mission types, airspace configurations, weather conditions, UAV designs, and payload categories. This approach yields a large-scale dataset, \texttt{UAVBench}, consisting of \textit{50000 validated and physically consistent UAV flight scenarios} that are semantically rich, safety-aware, and suitable for both training and evaluation.

  \item \textit{Multi-Stage Validation and Risk Labeling:} 
  We introduce a systematic validation pipeline that enforces geometric, physical, and safety constraints on all generated scenarios. Each scenario is further annotated with quantitative risk levels and categorical safety tags (e.g., \emph{Weather}, \emph{Navigation}, \emph{Energy}, \emph{Collision-Avoidance}) derived from the detected hazards and environmental severity, forming a reproducible and interpretable benchmark for risk-aware UAV autonomy.

  \item \textit{UAVBench\_MCQ (Structured Reasoning Benchmark):} 
  We present \texttt{UAVBench\_MCQ}, a reasoning-oriented extension of \texttt{UAVBench} containing \textit{50{,}000 multiple-choice questions (MCQs)} systematically derived from validated scenarios. Each MCQ follows a standardized JSON schema and belongs to one of ten reasoning styles—\emph{aerodynamics \& physics}, \emph{navigation \& path planning}, \emph{policy \& compliance}, \emph{environmental sensing}, \emph{multi-agent coordination}, \emph{cyber-physical security}, \emph{energy management}, \emph{ethical decision-making}, \emph{comparative systems}, and \emph{hybrid integrated reasoning}. The framework enforces grounded realism, structural completeness, and logical consistency to enable reproducible large-scale reasoning evaluation.

  \item \textit{Large-Scale LLM Evaluation:} 
  We benchmark thirty-two state-of-the-art large language models (LLMs)—including GPT-5, ChatGPT~4o, Gemini~2.5~Flash, DeepSeek~V3, Qwen3~235B, ERNIE~4.5~300B, and Mistral~Medium~3.1—on the \texttt{UAVBench\_MCQ} benchmark. 
  The evaluation spans ten reasoning styles covering physical, navigational, ethical, and hybrid cognitive dimensions of UAV autonomy, revealing strong performance in perception and policy reasoning but persistent challenges in multi-agent coordination, energy management, and ethics-aware decision-making.
\end{enumerate}

The remainder of this paper is structured as follows. 
Section~\ref{sec:related_work} reviews previous studies on LLM-driven autonomy, UAV simulation datasets, and reasoning benchmarks.  Section~\ref{sec:datasetgen} describes the construction of the \texttt{UAVBench} dataset, including its taxonomy-guided scenario generation, schema definition, and multi-stage validation and risk-labeling pipeline. 
Section~\ref{sec:exp_res} introduces \texttt{UAVBench\_MCQ}, a reasoning-focused extension that formalizes ten styles of cognitive and ethical reasoning for UAV systems.  Finally, Section~\ref{sec:conclusion} summarizes the main findings and discusses potential directions for future research in agentic and safety-aware UAV intelligence.

\section{Related Work}
\label{sec:related_work}

In recent years, substantial progress has been made toward developing benchmarks and platforms to evaluate embodied intelligence in unmanned aerial vehicles (UAVs). Various studies have proposed simulation frameworks, large-scale datasets, and evaluation methodologies, each addressing specific aspects, such as vision-language navigation, embodied cognition, and object-goal navigation. While these contributions have significantly advanced UAV autonomy, they exhibit notable differences in task definitions, experimental settings, and evaluation strategies.

\subsection{Vision-Language Navigation Platforms for UAVs}
Wang et al.~\cite{wang2024towards} introduce OpenUAV, a simulation platform designed to advance vision-language navigation (VLN) for UAVs. Unlike prior benchmarks that oversimplify aerial navigation using discrete actions, OpenUAV provides realistic environments, continuous six-degrees-of-freedom (6-DoF) flight control, and algorithmic support for trajectory generation. Using this platform, the authors construct the first large-scale dataset of realistic UAV VLN trajectories (over 12,000), enriched with human-annotated paths and GPT-4–generated navigation instructions. To address the challenges of aerial search tasks, they propose the UAV-Need-Help benchmark, which introduces assistant-guided navigation with varying levels of support. Finally, they develop a UAV navigation LLM that integrates multiview images, language instructions, and assistant guidance to produce hierarchical trajectories, achieving significant performance gains over baselines but still trailing behind human operators.

\subsection{Embodied Aerospace Intelligence}
Yao et al.~\cite{yao2024aeroverse} introduce AeroVerse, a benchmark designed to foster the development of embodied aerospace intelligence. The authors present AeroSimulator, a drone simulation platform that models realistic urban scenes using Unreal Engine and AirSim, alongside two large-scale pre-training datasets: AerialAgent-Ego10k (real-world drone image-text pairs) and CyberAgent-Ego500k (virtual image-text-pose alignment data). They also define, for the first time, five downstream tasks of the UAV agent: scene awareness, spatial reasoning, navigation exploration, task planning, and motion decision, and provide corresponding fine-tuning datasets (SkyAgent-Scene3k, SkyAgent-Reason3k, SkyAgent-Nav3k, SkyAgent-Plan3k, and SkyAgent-Act3k). To evaluate UAV agent performance, the authors propose SkyAgent-Eval, a GPT-4–based automated evaluation framework that complements traditional metrics such as BLEU and SPICE. Experimental results with multiple 2D/3D vision-language models highlight both the promise and limitations of existing approaches, underscoring the need for specialized aerospace embodied-world models.

\subsection{Benchmarks for UAV-Embodied Agents}
Guo et al.~\cite{guo2025bedi} propose BEDI, a framework to assess UAV-embodied agents (UAV-EAs). At its core is the Dynamic Chain-of-Embodied-Task paradigm, which models UAV behavior as a perception–decision–action loop and decomposes complex missions into measurable subtasks. Based on this paradigm, the benchmark defines five core skills—semantic perception, spatial perception, motion control, tool utilization, and task planning—and designs evaluation metrics for each. To ensure broad applicability, BEDI integrates both static real-world imagery and dynamic virtual environments (e.g., cargo delivery, firefighting, moving-target tracking), enabling agents to be tested under varied conditions. Importantly, it offers open interfaces for integrating custom UAV agents, promoting reproducibility and extensibility. Evaluations of several state-of-the-art vision-language models highlight their limitations in handling embodied UAV tasks, underscoring BEDI’s role in establishing a systematic, open, and scalable benchmark for UAV embodied intelligence.

\subsection{Object Goal Navigation in UAVs}
Xiao et al.~\cite{xiao2025uav} introduce UAV-ON, a benchmark dedicated to instance-level Object Goal Navigation (ObjectNav) in outdoor aerial settings. Unlike prior UAV vision-and-language navigation benchmarks that rely on detailed, step-by-step instructions, UAV-ON defines more than 11,000 navigation tasks using semantic goal instructions that describe object category, approximate size, and visual attributes. The benchmark features 14 high-fidelity outdoor environments created with Unreal Engine and AirSim, spanning urban, natural, and mixed-use regions, and includes 1,270 annotated target objects placed according to realistic co-occurrence patterns. UAV-ON employs physically grounded continuous controls rather than teleport-based movements, requiring agents to integrate perception, obstacle avoidance, and semantic reasoning for safe navigation. The authors evaluate three baselines: a random policy, a CLIP-based heuristic agent, and their Aerial ObjectNav Agent (AOA), a zero-shot framework leveraging multimodal LLM reasoning. Results reveal that while LLM-based approaches excel at semantic exploration, they struggle with precise stopping and safe trajectory execution, leading to high collision rates across all methods.

\subsection{Embodied Cognition in Urban Airspaces}
Zhao et al.~\cite{zhao2025urbanvideo} present a benchmark specifically designed to assess embodied cognition in motion within complex 3D urban environments. The benchmark introduces a novel task suite of 16 tasks across four categories—recall, perception, reasoning, and navigation—each designed to test the embodied capabilities of Video-LLMs. To support these tasks, the authors collected 1,547 embodied drone video clips from real cities in Guangdong Province and from two simulators (EmbodiedCity and AerialVLN), and generated over 5,200 multiple-choice questions (MCQs) using a hybrid pipeline that combines LLM-based generation, blind filtering, and human refinement. Seventeen Video-LLMs, both open-source and proprietary, were evaluated under zero-shot and fine-tuned settings. Results show that state-of-the-art models achieve only ~45\% accuracy, and that causal reasoning is strongly correlated with recall, perception, and planning. The study highlights the challenges of embodied intelligence in urban airspaces and demonstrates the potential of simulation-to-real transfer through fine-tuning.

\subsection{Comparative Analysis of UAV Benchmarks}
\label{subsec:benchmark_comparison}

Table~\ref{tab:uavbench_comparison_matrix} presents a comparative overview of recent UAV embodied-intelligence benchmarks, highlighting differences in design scope, reasoning coverage, and evaluation methodology. The analysis shows that prior work, such as \textit{OpenUAV}~\cite{wang2024towards} and \textit{AeroVerse}~\cite{yao2024aeroverse}, emphasizes physically realistic environments and vision-language navigation tasks but provides limited validation or risk modeling. Similarly, \textit{BEDI}~\cite{guo2025bedi} and \textit{UAV-ON}~\cite{xiao2025uav} advance embodied cognition and aerial navigation yet lack a unified schema or reasoning-based evaluation component. \textit{UrbanVideo-Bench}~\cite{zhao2025urbanvideo} focuses on video understanding and multimodal reasoning within urban contexts but remains narrow in mission diversity and lacks standardized cognitive evaluation. In contrast, UAVBench and its reasoning extension UAVBench\_MCQ provide comprehensive coverage across physical realism, validation and risk assessment, multi-domain reasoning, and structured evaluation. This unified design enables consistent, interpretable, and reproducible benchmarking of UAV intelligence across perception, planning, and decision-making dimensions.

Overall, these works represent significant progress toward advancing UAV autonomy by introducing specialized platforms, datasets, and evaluation frameworks. However, existing benchmarks still face several limitations. Many focus on narrow tasks, such as vision-language navigation or object-goal navigation, limiting their applicability to broader mission scenarios. Others remain heavily simulation-driven, limiting their ability to capture the full complexity of real-world aerial environments. In addition, task designs are often predefined and lack scalability, constraining the diversity of challenges UAV agents can face. Finally, the evaluation approaches remain fragmented, with limited emphasis on unified and systematic assessments of UAV intelligence across perception, reasoning, planning, and execution. These limitations highlight the need for more comprehensive, flexible, and realistic benchmarks—such as UAVBench and UAVBench\_MCQ—that can better support the development of next-generation UAV-embodied intelligence.

\begin{figure*}[htbp]
    \centering
    \includegraphics[width=1\textwidth]{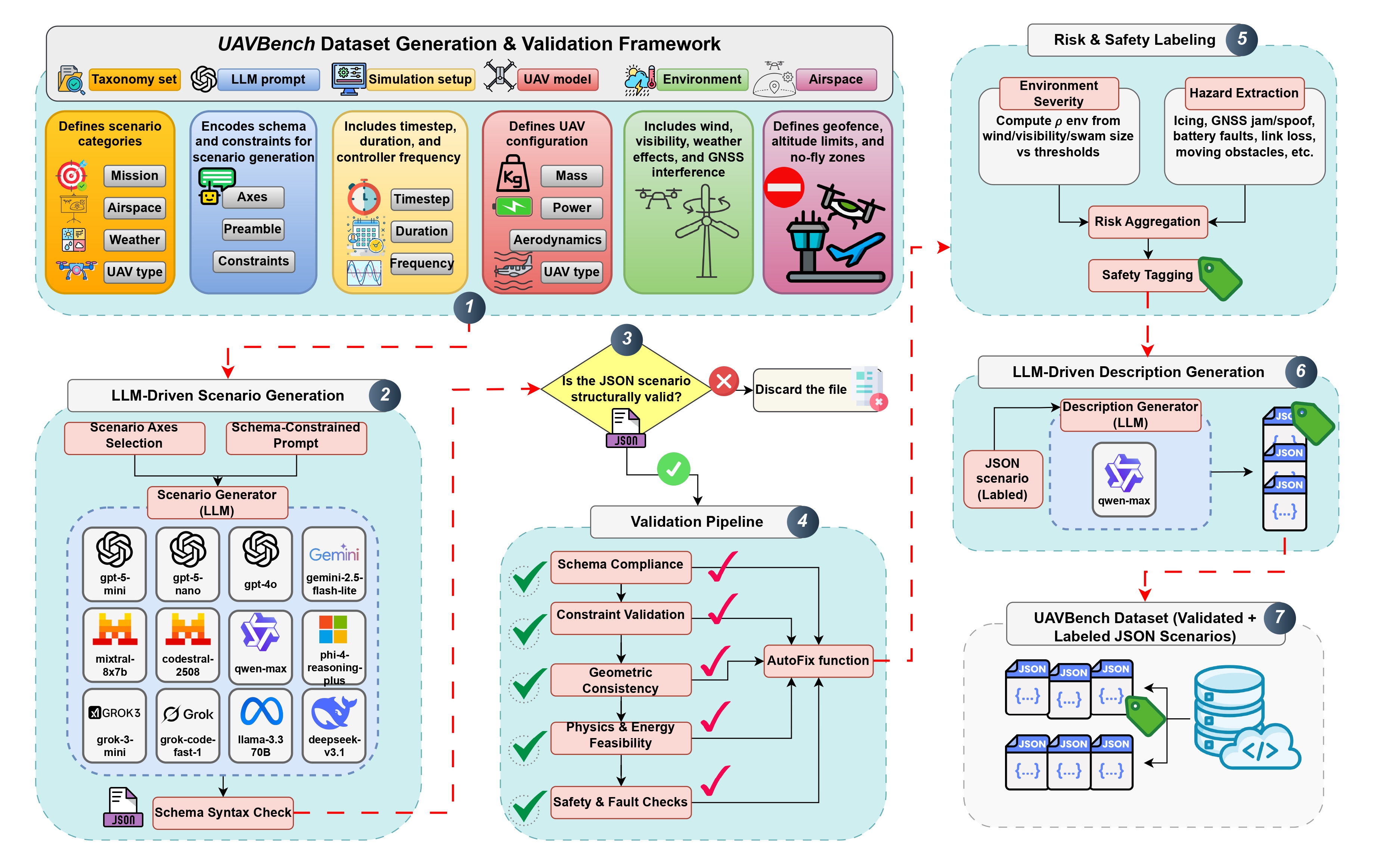} 
    \caption{UAVBench Dataset Generation, Validation, and Labeling Framework.}
    \label{fig:fig1}
\end{figure*}

\begin{table}[h!]
\centering
\scriptsize
\caption{Notation used in the dataset generation methodology.}
\label{tab:notation}
\renewcommand{\arraystretch}{0.7}
\rowcolors{2}{white}{cyanblue!70} 
\begin{tabular}{llp{4.5cm}}
\toprule
\textbf{Symbol} & \textbf{Block} & \textbf{Definition} \\
\midrule
$\mathcal{S}$ & Scenario & Full scenario tuple containing all blocks \\
\texttt{name} & Scenario & Scenario identifier string \\
\texttt{seed} & Scenario & Random seed for reproducibility \\
$\Sigma$ & Simulation & Simulation setup tuple \\
$\Delta t$ & Simulation & Integration time step (s) \\
$N$ & Simulation & Number of simulation steps \\
$f_c$ & Simulation & Controller update frequency (Hz) \\
$T$ & Simulation & Total mission duration ($T=N \cdot \Delta t$) \\
\midrule
$\mathcal{U}$ & UAV & UAV configuration block \\
$\tau$ & UAV & UAV type (e.g., quadrotor, fixed-wing) \\
$m$ & UAV & Mass (kg) \\
$E_b$ & UAV & Battery energy (Wh) \\
$V_f$ & UAV & Fuel volume (L) \\
$\xi$ & UAV & Energy source type (battery, fuel, hybrid) \\
$v_{\max}$ & UAV & Maximum velocity (m/s) \\
$\phi_{\max}$ & UAV & Maximum tilt angle (deg) \\
$r$ & UAV & Reserve energy fraction \\
$P(v,\dot{\mathbf{u}})$ & UAV & Power consumption model \\
$P_h$ & UAV & Hover power (W) \\
$k_d$ & UAV & Drag coefficient \\
$k_m$ & UAV & Maneuver coefficient \\
$A_d$ & UAV & Rotor disk area (m$^2$) \\
$C_D$ & UAV & Drag coefficient (fixed-wing) \\
$AR$ & UAV & Aspect ratio \\
$e$ & UAV & Oswald efficiency factor \\
$S$ & UAV & Wing area (m$^2$) \\
$C_{L,\max}$ & UAV & Maximum lift coefficient \\
$v_{\text{stall}}$ & UAV & Stall speed (m/s) \\
\midrule
$\mathcal{P}$ & Payload & Set of payload elements \\
$p_i$ & Payload & Payload element $i$ \\
$t_i$ & Payload & Payload type (e.g., lidar, camera, repeater) \\
$m_i$ & Payload & Mass (kg) \\
$P_i$ & Payload & Power (W) \\
$C_{dA,i}$ & Payload & Drag–area coefficient (m$^2$) \\
$\mu_i$ & Payload & Mount position \\
$\mathcal{D}_i$ & Payload & Data parameters \\
$\mathcal{O}_i$ & Payload & Operating parameters \\
$\mathcal{C}_i$ & Payload & Constraints \\
$m_{\text{tot}}$ & Payload & Total mass ($m + \sum_i m_i$) \\
$P_{\text{payload}}$ & Payload & Total payload power ($\sum_i P_i$) \\
$D_{\text{payload}}(v)$ & Payload & Payload drag term \\

\midrule
$\mathcal{E}$ & Environment & Environment block \\
$\mathcal{E}_w$ & Environment & Weather tuple \\
$w$ & Environment & Wind speed (m/s) \\
$\psi$ & Environment & Wind direction (deg) \\
$g$ & Environment & Gust amplitude (m/s) \\
$\gamma$ & Environment & Visibility condition \\
$\Phi$ & Environment & Atmospheric phenomena (hail, icing, etc.) \\
$J_{\text{GNSS}}$ & Environment & GNSS jamming power (dBm) \\
\midrule
$\mathcal{A}$ & Airspace & Airspace definition \\
$h_{\min}, h_{\max}$ & Airspace & Minimum and maximum altitude (m) \\
$\mathcal{P}_\ell$ & Airspace & Polygonal geofence region \\
\midrule
$\mathcal{X}_0$ & Spawn & Initial UAV spawn state \\
\midrule
$\mathcal{M}$ & Mission & Mission block \\
$\sigma$ & Mission & Mission type \\
$\mathcal{WP}$ & Mission & Set of waypoints \\
$\kappa$ & Mission & Path pattern (grid, corridor, orbit) \\
$r_\ell$ & Mission & Loiter radius (m) \\
$B$ & Mission & Time budget (s) \\
$\rho_{\text{rw}}$ & Mission & Runway requirement flag \\
$\Upsilon$ & Mission & VTOL transition profile \\
\midrule
$\mathcal{T}$ & Entities & Background traffic \\
$\mathcal{O}$ & Entities & Moving obstacles \\
$\mathcal{W}$ & Entities & Swarm teammates \\
$d_{\min}$ & Entities & Minimum inter-UAV separation (m) \\
\midrule
$\mathcal{C}$ & Control & Control action set \\
$\mathsf{CtrlOK}(\tau,\mathcal{A})$ & Control & Predicate: UAV type $\tau$ has valid control set \\
\midrule
$\mathcal{D}$ & Safety & Safety thresholds \\
$d_{\text{sep}}$ & Safety & Required separation distance (m) \\
$\text{TTC}_{\min}$ & Safety & Minimum time-to-collision (s) \\
\midrule
$\mathcal{F}$ & Faults & Fault injection block \\
$(t_i,\varphi_i,\Delta t_i,s_i)$ & Faults & Fault event tuple: start time, type, duration, severity \\
\midrule
$\mathcal{L}$ & Comms & Communication constraints (uplink, downlink, signal strength) \\
\midrule
$\theta$ & Prompt & Axis tuple for LLM prompt \\
$s,a,w,u,\nu$ & Prompt & Scenario, airspace, weather, UAV type, and nonce \\
$\Pi(\mathbb{S},\mathbb{C};\theta)$ & Prompt & LLM prompt construction function \\
\midrule
$\rho(S)$ & Risk & Risk score of a scenario $S$ \\
$\sigma(S)$ & Risk & Safety category label of a scenario $S$ \\
\bottomrule
\end{tabular}
\end{table}

\section{UAVBench - Dataset Generation Methodology}
\label{sec:datasetgen}

Designing a benchmark for UAV autonomy requires scenarios that are both systematically diverse and scientifically rigorous.  To achieve this, we formalize each scenario as a structured tuple that captures simulation dynamics, UAV configuration, environmental disturbances, mission objectives, and safety constraints. Table~\ref{tab:notation} summarizes the core mathematical notation used throughout this section.

Fig. \ref{fig:fig1} presents the complete \texttt{UAVBench} pipeline, illustrating the structured process for generating, validating, and labeling UAV scenarios. The framework begins with taxonomy-guided LLM scenario generation, followed by multi-stage validation that ensures schema compliance, geometric and physical feasibility, and safety consistency. Validated scenarios are then risk-scored and safety-tagged before being passed to an auxiliary LLM module that produces concise human-readable descriptions. The resulting dataset—composed of validated and labeled JSON files—is ready for benchmarking and simulation of agentic AI systems in diverse UAV mission contexts.

\subsection{Scenario Schema Design}
\label{subsec:schema}

The benchmark relies on a structured schema that ensures that each scenario is both syntactically valid and physically realistic. To make this explicit, we formalize the schema using mathematical notation and provide detailed definitions of all symbols. A scenario is represented as follows.
\begin{equation}
\mathcal{S} = \langle \texttt{name}, \texttt{seed}, \Sigma, \mathcal{U}, \mathcal{E}, \mathcal{A},
\mathcal{X}_0, \mathcal{M}, \mathcal{T}, \mathcal{O}, \mathcal{W}, \mathcal{C}, \mathcal{D}, \mathcal{F}, \mathcal{L} \rangle ,
\end{equation}
where $\Sigma$ defines the simulation parameters, $\mathcal{U}$ the UAV model, $\mathcal{E}$ the environment, $\mathcal{A}$ the airspace, $\mathcal{X}_0$ the initial spawn state, $\mathcal{M}$ the mission, $\mathcal{T}$ background traffic, $\mathcal{O}$ obstacles, $\mathcal{W}$ swarm teammates, $\mathcal{C}$ the control space, $\mathcal{D}$ safety thresholds, $\mathcal{F}$ injection faults and $\mathcal{L}$ communication constraints. The following subsections provide detailed definitions of each block.

\subsubsection{Simulation Setup}
The simulation setup controls the temporal structure of each scenario. It is defined as:
\begin{equation}
\Sigma = \langle \Delta t, N, f_c \rangle ,
\end{equation}
where $\Delta t$ is the integration time step (s), $N$ the number of discrete simulation steps, and $f_c$ the controller update frequency (Hz). The constraints are:
\begin{equation}
\Delta t \in [0.01,0.05], \quad N \geq 600, \quad f_c \geq 10 .
\end{equation}
The duration of the mission $T$ is then:
\begin{equation}
T = N \cdot \Delta t .
\end{equation}

These conditions ensure that each simulation runs for a non-trivial duration and at a temporal resolution appropriate for UAV dynamics. If $\Delta t$ is too small, computational cost becomes excessive, while if it is too large, important dynamics may be missed. Similarly, $N \geq 600$ prevents trivial short-hop scenarios and aligns the dataset with real-world UAV missions that typically last several minutes. The controller frequency bound reflects the operational limits of autopilots such as PX4 and ArduPilot, grounding the benchmark in practical system architectures.

\subsubsection{UAV Configuration and Propulsion}
The UAV block specifies the physical and energetic configuration:
\begin{equation}
\mathcal{U} = \langle \tau, m, E_b, V_f, \xi, v_{\max}, \phi_{\max}, r, \mathcal{B}, \mathcal{R}, \mathcal{A}_f, \mathcal{Z}, \mathcal{P} \rangle .
\end{equation}
Here, $\tau$ is the UAV type, $m$ mass (kg), $E_b$ battery energy (Wh), $V_f$ fuel volume (L), $\xi$ energy source (\texttt{battery}/\texttt{fuel}/\texttt{hybrid}), $v_{\max}$ maximum velocity (m/s), $\phi_{\max}$ maximum tilt (deg), and $r$ reserved energy fraction. The sub-blocks are: $\mathcal{B}$ battery model (e.g., hover power and coefficients), $\mathcal{R}$ rotorcraft parameters (e.g., rotor count, disk area), $\mathcal{A}_f$ fixed-wing/forward-flight aerodynamics, $\mathcal{Z}$ sensors (renamed to avoid conflict with $\mathcal{S}$), and $\mathcal{P}$ payload.

Energy consumption is modeled as:
\begin{equation}
P(v,\dot{\mathbf{u}}) = P_h + k_d v^3 + k_m \|\dot{\mathbf{u}}\|_2 ,
\end{equation}
where $P_h$ is hover power (W), $k_d$ drag coefficient, $v$ velocity (m/s), $k_m$ maneuver coefficient, and $\|\dot{\mathbf{u}}\|_2$ the control-rate magnitude. The feasibility condition is:
\begin{equation}
\sum_{k=0}^{N-1} P_k \,\Delta t \leq (1-r)\, E_b \cdot 3600 ,
\end{equation}
where $P_k$ is the discrete power at step $k$ and the factor $3600$ converts Wh to joules.

Rotorcraft require rotor disk checks:
\begin{equation}
P_h \approx c_\eta \frac{m^{3/2}}{\sqrt{A_d}} ,
\end{equation}
where $c_\eta$ aggregates propulsive efficiency factors and $A_d$ is rotor disk area (m$^2$). For fixed-wing UAVs, aerodynamics are:
\begin{equation}
C_D = C_{D0} + \frac{1}{\pi\, AR\, e}\, C_L^2 , \qquad
v_{\text{stall}} \approx \sqrt{\frac{2 m g}{\rho\, S\, C_{L,\max}}}\,,
\end{equation}
with $C_{D0}$ parasitic drag, $AR$ aspect ratio, $e$ Oswald factor, $\rho$ air density, $S$ wing area, and $C_{L,\max}$ maximum lift coefficient.

This block ensures UAVs are not arbitrary numerical constructs but physically realistic platforms. It encodes first-order aerodynamic relationships, preventing, for example, “impossible” rotorcraft from carrying large payloads with tiny rotors. By embedding both rotorcraft and fixed-wing models, the schema spans the full design space of UAVs used in research and industry. The inclusion of reserve fractions aligns with operational safety practices, which require UAVs to always retain energy for contingencies.

\subsubsection{Payload Taxonomy and Examples}

UAVBench introduces a comprehensive payload taxonomy that reflects the diversity of sensing, communication, delivery, industrial, and defense systems employed across modern unmanned aerial vehicle missions.  
Each payload is represented as
\begin{equation}
p_i = \langle t_i, m_i, P_i, C_{dA,i}, \mu_i \rangle ,
\end{equation}
where $t_i$ denotes the canonical payload type (for example, lidar or thermal camera), $m_i$ is the payload mass (kg), $P_i$ is the electrical power requirement (W), $C_{dA,i}$ is the projected drag–area coefficient, and $\mu_i$ is the mounting position (belly, nose, top, wing, gimbal, bay, or tether).  
This standardized representation ensures that each payload contributes realistically to the total UAV mass, power consumption, and aerodynamic profile.

The \texttt{UAVBench} taxonomy enumerates more than 200 canonical payload types, grouped into over 30 functional categories spanning the civil, scientific, and defense domains.  
The taxonomy covers a wide range of mission contexts:

\begin{itemize}[leftmargin=1.2em]
    \item Imaging and sensing: optical, thermal, multispectral, hyperspectral, and advanced imaging such as lidar or synthetic-aperture radar.
    \item Communication and networking: radio relay, cellular base stations, satellite communication terminals, and ad-hoc mesh nodes.
    \item Industrial and environmental monitoring: methane and gas detection, pipeline inspection, and ground-penetrating radar.
    \item Public safety and emergency response: search-and-rescue thermal cameras, fire-mapping systems, and emergency beacons.
    \item Delivery and logistics: parcel and medical carriers, aerial drop systems, and life-raft deployment.
    \item Scientific and environmental research: radiation detectors, atmospheric sensors, and biosensors.
    \item Agricultural and ecological monitoring: multispectral and NDVI cameras, soil-moisture and crop-health sensors.
    \item Military and defense applications: electro-optical reconnaissance, laser designators, electronic-warfare systems, and CBRN detectors.
\end{itemize}

Table~\ref{tab:payload_examples} presents representative payloads from these categories, illustrating the breadth of the \texttt{UAVBench} taxonomy and typical physical parameters used in simulation.

\begin{table*}[h!]
\centering
\renewcommand{\arraystretch}{1.05}
\caption{Representative payload examples in \texttt{UAVBench} (excerpt from over 200 canonical types).}
\label{tab:payload_examples}
\renewcommand{\arraystretch}{0.7}
\rowcolors{2}{white}{cyanblue!70}
\begin{tabular}{l l l r r l}
\toprule
\textbf{Category} & \textbf{Type (canonical)} & \textbf{Mount} & \textbf{Mass [kg]} & \textbf{Power [W]} & \textbf{Primary use} \\
\midrule
Imaging optical & Gimbaled camera & Gimbal & 0.35 & 6 & Structural inspection \\
Imaging thermal & Thermal camera (LWIR) & Nose & 0.30 & 5 & Night operations or SAR \\
Imaging multispectral & Multispectral camera & Wing & 0.45 & 8 & Vegetation and crop analysis \\
Imaging advanced & Lidar sensor & Belly & 1.20 & 18 & Terrain and infrastructure mapping \\
Communication relay & Radio repeater & Bay & 0.60 & 10 & Network extension \\
Communication link & Satellite communication terminal & Top & 1.10 & 25 & BVLOS operation \\
Industrial monitoring & Methane detection sensor & Belly & 0.80 & 12 & Oil and gas inspection \\
Industrial sensing & Ground-penetrating radar & Belly & 2.50 & 25 & Subsurface mapping \\
Public safety & SAR thermal camera & Gimbal & 0.35 & 6 & Search and rescue \\
Emergency response & Fire-thermal mapping system & Nose & 0.50 & 10 & Fire surveillance \\
Delivery & Parcel delivery system & Bay/tether & 1.50 & 3 & Urban logistics \\
Medical delivery & Medical supply carrier & Bay & 2.00 & 4 & Emergency medicine transport \\
Scientific & Gamma-ray spectrometer & Bay & 1.00 & 9 & Radiation monitoring \\
Agricultural & NDVI camera & Wing & 0.40 & 7 & Crop-health mapping \\
Ecological & Soil-moisture sensor & Belly & 0.35 & 4 & Environmental survey \\
Military ISR & Electro-optical reconnaissance system & Nose & 1.80 & 20 & Intelligence and surveillance \\
Military EW & Electronic-warfare jammer & Bay & 2.20 & 30 & Signal denial and counter-UAS \\
\bottomrule
\end{tabular}
\end{table*}

The inclusion of such a broad and standardized payload taxonomy enables \texttt{UAVBench} to generate mission scenarios with realistic, heterogeneous configurations—ranging from lightweight electro-optical cameras on micro-UAVs to multi-sensor payload suites on large fixed-wing platforms.  
By integrating payload mass, power, and aerodynamic drag directly into the simulation model, \texttt{UAVBench} enables reproducible, physically consistent benchmarking of autonomy, mission planning, and energy-aware flight control across both civilian and defense contexts.

\subsubsection{Environment and Airspace}
The environment $\mathcal{E}$ specifies weather and disturbances via the weather tuple
\begin{equation}
\mathcal{E}_w = \langle w, \psi, g, \gamma, \Phi \rangle ,
\end{equation}
where $w$ is wind speed (m/s), $\psi$ wind direction (deg), $g$ gust amplitude (m/s), $\gamma$ visibility condition (categorical), and $\Phi$ atmospheric phenomena (e.g., icing, sandstorm). Optional electromagnetic effects include GNSS multipath, jamming power $J_{\text{GNSS}}$ (dBm), and general EM interference.

Airspace $\mathcal{A}$ encodes altitude and lateral constraints. With $h_{\max} > h_{\min}$, vertical limits are enforced. Waypoints $\vect{w}_i$ must lie inside geofences:
\begin{equation}
\vect{w}_i \in \bigcup_\ell \mathcal{P}_\ell \,,
\end{equation}
where $\mathcal{P}_\ell$ are polygonal regions. No-fly zones are cylinders (static or dynamic), while runways are included for fixed-wing UAVs.

This block introduces environmental realism by constraining UAVs to atmospheric and regulatory conditions. Limiting wind speeds reflects the upper thresholds of UAV flight envelopes, while visibility levels capture operational categories such as VFR/IFR. Including electromagnetic effects such as GNSS jamming allows scenarios to simulate contested environments, aligning the benchmark with security and resilience studies. Altogether, this block integrates physical, meteorological, and regulatory realism.

\subsubsection{Mission and Entities}
The mission block $\mathcal{M}$ specifies task objectives:
\begin{equation}
\mathcal{M} = \langle \sigma, \mathcal{WP}, \kappa, r_\ell, B, \rho_{\text{rw}}, \Upsilon \rangle ,
\end{equation}
where $\sigma$ is mission type, $\mathcal{WP}$ the set of waypoints ($3 \leq |\mathcal{WP}| \leq 6$), $\kappa$ path pattern, $r_\ell$ loiter radius (m), $B$ time budget (s), $\rho_{\text{rw}}$ runway requirement flag, and $\Upsilon$ VTOL transition profile.

External entities enrich realism. Traffic $\mathcal{T}$ defines background UAVs, $\mathcal{O}$ moving obstacles, and $\mathcal{W}$ swarms. Swarm separation is enforced by:
\begin{equation}
\|\vect{x}_a(t) - \vect{x}_b(t)\|_2 \geq d_{\min}, \quad \forall t \in [0,T] .
\end{equation}

This block ensures that missions are neither trivial nor overly complex. Limiting waypoints to between three and six captures real-world planning tasks such as corridor inspections or grid-based surveys. The time budget introduces trade-offs between task completion and energy limits, mimicking operational decision-making. Including swarms and moving obstacles tests cooperative and reactive autonomy, which are critical features of next-generation UAV systems operating in dense airspaces.

\subsubsection{Control, Safety, and Faults}
The control block $\mathcal{C}$ specifies UAV action sets, either discrete or continuous. A predicate $\mathsf{CtrlOK}(\tau,\mathcal{A})$ ensures each UAV has sufficient degrees of freedom to remain flyable. For example, a fixed-wing UAV must support throttle, pitch, roll, and yaw.

Safety thresholds are captured as:
\begin{equation}
\mathcal{D} = \langle d_{\text{sep}}, \text{TTC}_{\min} \rangle ,
\end{equation}
with violations defined as:
\begin{equation}
\min_{a \neq b} \|\vect{x}_a(t)-\vect{x}_b(t)\|_2 < d_{\text{sep}}
\quad \text{or} \quad
\min_{a \neq b} \text{TTC}_{ab}(t) < \text{TTC}_{\min}.
\end{equation}

Faults are modeled as tuples $(t_i,\varphi_i,\Delta t_i,s_i)$: $t_i$ start time (s), $\varphi_i$ type (e.g., motor failure, GNSS jam), $\Delta t_i$ duration (s), and $s_i$ severity. Communication constraints $\mathcal{L}$ define uplink/downlink availability and signal thresholds.

This block allows UAVs to be benchmarked not only in nominal conditions but also in degraded environments. Safety thresholds align with the Unmanned Traffic Management (UTM) literature \cite{hamissi2023survey}, ensuring comparability with regulatory concepts. The inclusion of fault injection makes the benchmark suitable for resilience testing, capturing how autonomy responds to failures such as GNSS denial or sensor corruption. This elevates the schema beyond static mission planning, positioning it as a tool for robustness evaluation.

\begin{figure*}[h!]
  \centering
  \includegraphics[width=1\textwidth]{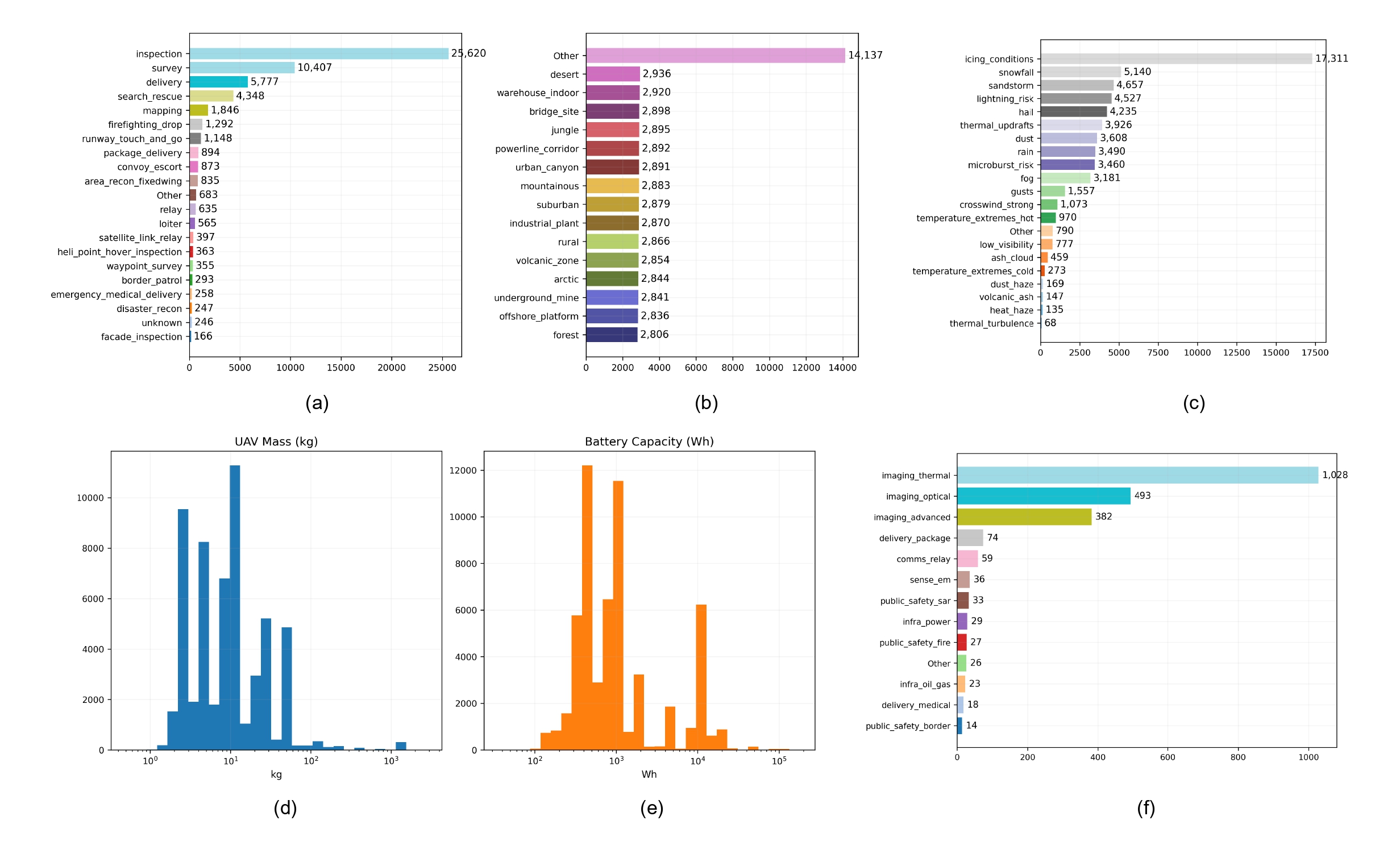} 
  \caption{
  Overview of \texttt{UAVBench} dataset composition and UAV design characteristics.
  (a) Mission types illustrating the diversity of operational scenarios;
  (b) Airspace types showing the range of environmental contexts;
  (c) Weather phenomena highlighting atmospheric complexity;
  (d) UAV mass distribution indicating variability in platform sizes;
  (e) Battery capacity distribution reflecting energy endurance profiles; and
  (f) Payload category frequencies summarizing the variety of onboard sensors and mission payloads.}
  \label{fig:uavbench_overview}
\end{figure*}

\subsection{Taxonomies and Prompt Design for LLM-Based Scenario Generation}
\label{subsec:taxonomies-prompt}

To ensure that LLM-generated scenarios are both diverse and operationally realistic, we introduce structured taxonomies for scenarios, airspaces, weather, and UAV types. Each taxonomy organizes discrete tokens into meaningful categories, which are then embedded in the LLM prompt along with explicit constraints. This combination allows us to balance flexibility and validity in scenario generation. Fig.~\ref{fig:uavbench_overview} presents an overview of the \texttt{UAVBench} dataset composition and UAV platform characteristics. The distributions illustrate the dataset’s diversity across mission types, airspace configurations, and weather conditions, highlighting its coverage of complex and realistic operational contexts. The UAV-specific statistics, including mass and battery capacity, reveal a wide range of vehicle sizes and endurance profiles, while payload diversity reflects the presence of multiple sensing modalities, including optical, thermal, and radar units. Collectively, these aspects demonstrate UAVBench’s suitability for benchmarking perception, autonomy, and risk-aware decision-making algorithms under heterogeneous aerial scenarios.

\subsubsection{Scenario Taxonomy}
Scenarios capture the mission-level intent of UAV operations. We organize them into categories such as inspection, delivery, reconnaissance, search and rescue, training, swarm coordination, safety-critical events, fire/hazmat, and maritime/offshore. Formally, we define a mapping
\begin{equation}
\mathcal{C}_S : c \mapsto \{s_1, s_2, \dots, s_n\},
\end{equation}
where $c$ is a scenario category (e.g., \textit{inspection}) and $\{s_i\}$ is the set of scenario tokens in that category. This organization enables us to ensure that generated missions span different operational classes, from bridge inspections with multirotors to BVLOS mountain ridge flights with fixed-wing aircraft.

\subsubsection{Airspace Taxonomy}
Airspaces represent the environments in which UAVs operate. We partition them into four broad groups: urban, natural terrain, infrastructure corridors, and special constrained zones. This can be represented as

\begin{equation}
\mathcal{C}_A : a \mapsto \{e_1, e_2, \dots, e_m\},
\end{equation}

where $a$ is an airspace category and $\{e_j\}$ are the specific environments (e.g., \texttt{urban\_canyon}, \texttt{desert}, \texttt{underground\_mine}). This structure enables analysis of whether generated scenarios sufficiently explore both routine and extreme environments and ensures that waypoints remain consistent with airspace constraints.

\subsubsection{Weather Taxonomy and Severity}
Weather conditions influence both mission safety and UAV performance. We group tokens into precipitation, wind, visibility, icing/temperature extremes, electrical risks, and clear conditions. In addition, we define a severity function
\begin{equation}
\sigma_w : w \mapsto \{0,1,2,3,4\},
\end{equation}
where $w$ is a weather token and $\sigma_w(w)$ gives its ordinal severity score (e.g., $\sigma_w(\texttt{clear})=0$, $\sigma_w(\texttt{rain})=1$, $\sigma_w(\texttt{icing\_conditions})=4$). This abstraction prevents the generation of implausible or unsafe combinations while preserving diversity across environmental conditions.

\subsubsection{UAV Type Taxonomy}
UAV platforms are classified into multirotors, rotorcraft, fixed-wing/gliders, and hybrid concepts. We model this as
\begin{equation}
\mathcal{C}_U : u \mapsto \{t_1, t_2, \dots, t_k\},
\end{equation}
where $u$ is a UAV family (e.g., \textit{multirotors}) and $\{t_k\}$ are the concrete vehicle types (e.g., \texttt{quadrotor}, \texttt{hexacopter}). This taxonomy captures fundamental differences in dynamics and operational envelopes, ensuring that fixed-wing missions require runways, while rotorcraft missions can operate in constrained urban areas.

\subsubsection{Prompt Integration}
Finally, we integrate the taxonomies into the LLM prompt to guide scenario generation. Each prompt selects one element from each taxonomy and combines it with schema-level constraints. We formalize the axis tuple as
\begin{equation}
\theta = \langle s, a, w, u, \nu \rangle,
\end{equation}
where $s \in \mathcal{C}_S$, $a \in \mathcal{C}_A$, $w \in \mathcal{C}_\mathcal{E}$, $u \in \mathcal{C}_U$, and $\nu$ is a random nonce that increases diversity. The prompt then embeds $\theta$ along with explicit constraints such as simulation duration, waypoint counts, or runway requirements. This structured design ensures that the LLM produces JSON objects that are not only valid but also operationally meaningful.

\subsubsection{LLM Prompt Mechanics}
We compose a single instruction that binds the taxonomy choices to the JSON schema and explicit guardrails. Let the prompt be $\Pi(\mathbb{S},\mathbb{C};\theta)$, where $\mathbb{S}$ is the verbatim JSON schema, $\mathbb{C}$ is the constraint bullet list, and $\theta=\langle s,a,w,u,\nu\rangle$ are the “axes’’: scenario type $s$, airspace $a$, weather $w$, UAV type $u$, and a nonce $\nu$ to decorrelate outputs. The function first samples $\theta$ from the canonicalized taxonomies, then interpolates $u$ into type-conditional constraints (e.g., requiring \texttt{rotorcraft} and/or \texttt{aero} blocks), and finally concatenates \emph{(i)} a strict preamble (“JSON only”), \emph{(ii)} the axes header, \emph{(iii)} the full schema $\mathbb{S}$, and \emph{(iv)} the constraints $\mathbb{C}$. This “specification-by-example” design narrows the LLM’s search space at generation time, significantly reducing out-of-range values (e.g., \texttt{dt}) and structural errors (e.g., missing geofence or runway). The nonce $\nu$ preserves diversity across repeated calls without weakening the constraints or the schema.

\subsection{Validation Pipeline}

The validation pipeline ensures that each scenario produced by the LLM is not only syntactically valid but also semantically consistent with the schema and physically plausible. This step is essential because large language models may generate well-structured outputs that nonetheless contain hidden inconsistencies or unrealistic mission details. By incorporating a multi-stage validation process, \texttt{UAVBench} transforms raw generative outputs into reliable and reusable benchmark assets. The process can be represented as Algorithm~\ref{alg:validation}.

\begin{algorithm}[h!]
\caption{Validation Pipeline for LLM-Generated Scenarios}
\label{alg:validation}
\KwIn{Scenario $S$ generated by LLM}
\KwOut{Validity flag $valid \in \{\texttt{true},\texttt{false}\}$}

\tcp{Schema Compliance}
Check that all required keys 
$K=\{\texttt{name}, \texttt{seed}, \texttt{sim}, \texttt{uav}, \texttt{environment}, \newline \texttt{airspace}, \texttt{spawn}, \texttt{mission}\}$ 
are present and well-typed\;
\eIf{$\exists k \in K : k \notin dom(S)$}{\Return false}{continue}

\tcp{Constraint Validation}
Let $s \leftarrow S.\texttt{mission.type}$ (where $s \in \mathcal{C}_S$)\;
Check that UAV type, airspace, and weather satisfy $\mathcal{C}(s)$\;
\eIf{$S \not\models \mathcal{C}(s)$}{\Return false}{continue}

\tcp{Geometric Consistency}
For each waypoint $w=(x,y,z)$ in $S$, verify $(x,y) \in G$ and $z_{\min} \leq z \leq z_{\max}$\;
If any waypoint violates constraint: \Return false\;

\tcp{Safety and Fault Checks}
For all UAV pairs $(i,j)$ compute distance $d_{ij}$ and time-to-collision $\tau_{ij}$\;
Check $d_{ij} \geq d_{\min}$ and $\tau_{ij} \geq \tau_{\min}$\;
Validate each fault event $(t_i,\varphi_i,s_i)$: $t_i \geq 0$, $s_i \leq s_{\max}$\;
If violations found: \Return false\;

\Return true\;
\end{algorithm}

\paragraph{Discussion.}
Algorithm~\ref{alg:validation} formalizes the multi-stage filtering process that ensures each generated scenario is valid. The input $S$ denotes a candidate scenario, encoded as a structured mapping according to the schema $\mathcal{S}$ defined in Section~\ref{subsec:schema}. The output is a Boolean validity flag $valid \in \{\texttt{true},\texttt{false}\}$ indicating whether the scenario is accepted into the benchmark.

The first stage checks \emph{schema compliance}. The set $K$ contains all mandatory keys that must be present in every scenario, namely \texttt{name}, \texttt{seed}, \texttt{sim}, \texttt{uav}, \texttt{environment}, \texttt{airspace}, \texttt{spawn}, and \texttt{mission}. If any key $k \in K$ is missing from the domain $\text{dom}(S)$ of the scenario, the scenario is immediately rejected. This guarantees structural completeness and prevents parsing errors in downstream simulation.

The second stage enforces \emph{constraint validation}. Let $s \in \mathcal{C}_S$ denote the mission type chosen in the scenario (e.g., \texttt{inspection}, \texttt{delivery}, \texttt{search\_and\_rescue}). For each mission type $s$, there exists a corresponding set of operational constraints $\mathcal{C}(s)$ that specifies which UAV types, airspace conditions, and weather profiles are admissible. The predicate $S \models \mathcal{C}(s)$ indicates that the scenario satisfies these rules. This step prevents illogical or unsafe pairings, such as fixed-wing aircraft operating underground or rotorcraft attempting satellite-relay missions.

The third stage verifies \emph{geometric consistency}. Each waypoint $w=(x,y,z)$ is defined in three-dimensional space, where $(x,y)$ are ground-plane coordinates and $z$ is altitude above ground. The polygonal geofence $G \subset \mathbb{R}^2$ defines lateral bounds, while the altitude interval $[z_{\min},z_{\max}]$ defines vertical limits. The condition $(x,y) \in G$ and $z_{\min} \leq z \leq z_{\max}$ ensures that all waypoints, spawn points, and landing sites remain within authorized operational boundaries. This prevents violations such as waypoints outside the geofence or below ground level, which would render scenarios infeasible in simulation.

The final stage applies \emph{safety and fault checks}. For every pair of UAVs $(i,j)$, the Euclidean distance $d_{ij}$ and time-to-collision $\tau_{ij}$ are computed. These must satisfy $d_{ij} \geq d_{\min}$ and $\tau_{ij} \geq \tau_{\min}$, where $d_{\min}$ and $\tau_{\min}$ are safety thresholds defined in the schema block $\mathcal{D}$. Fault events are represented as tuples $(t_i,\varphi_i,s_i)$, where $t_i$ is the event start time, $\varphi_i$ the fault type (e.g., motor failure, GNSS jam), and $s_i$ the severity. The validator ensures that $t_i \geq 0$ and $s_i \leq s_{\max}$, thereby excluding unrealistic cases such as an instantaneous catastrophic fault at mission start or a severity outside calibrated limits. 

In summary, the validation pipeline acts as a layered filter that combines schema-level checks ($K$), operational constraints ($\mathcal{C}(s)$), geometric feasibility ($G$, $[z_{\min},z_{\max}]$), and safety thresholds ($d_{\min}, \tau_{\min}, s_{\max}$). This guarantees that only structurally complete, logically coherent, spatially consistent, and operationally safe scenarios are admitted to the dataset. Such rigor is crucial for benchmarking agentic AI systems, since it ensures that evaluation results reflect meaningful performance rather than artifacts of poorly constructed scenarios.

\subsection{Risk \& Safety Labeling}

The labeling process assigns each validated scenario a discrete risk level and a categorical safety tag. This is implemented as a deterministic algorithm that combines hazard detection, environmental conditions, and mission context into a unified scoring procedure.

\begin{algorithm}[h!]
\caption{Risk and Safety Labeling Procedure}
\label{alg:risk}
\KwIn{Scenario $S = (H, E, M)$, where $H$ = hazards, $E$ = environment, $M$ = mission parameters}
\KwOut{Risk level $\rho(S) \in \{0,1,2,3\}$, Safety category $\sigma(S) \in \Sigma$}

$F(S) \leftarrow$ detect hazards from $H$ (e.g., icing, GNSS jamming, battery failure)\;
$v_{\text{wind}}, \gamma_{\text{vis}}, n_{\text{swarm}} \leftarrow$ extract environmental features from $E$\;

\eIf{$F(S) \neq \emptyset$}{
    $\rho_{\text{hazards}}(S) \leftarrow$ max severity of hazards in $F(S)$\;
}{
    $\rho_{\text{hazards}}(S) \leftarrow 0$\;
}

$\rho_{\text{env}}(S) \leftarrow$ severity score based on thresholds: \\
\Indp
    if $v_{\text{wind}} > v_{\text{th}}$ then add penalty\;  
    if $\gamma_{\text{vis}} = \text{poor}$ then add penalty\;  
    if $n_{\text{swarm}} > n_{\text{th}}$ then add penalty\;  
\Indm

$\rho(S) \leftarrow \max(\rho_{\text{hazards}}(S), \rho_{\text{env}}(S))$\;

$\sigma(S) \leftarrow$ assign category in $\Sigma$ based on dominant hazard \\
\Indp
(e.g., Weather, Navigation, Energy, Collision-Avoidance)\;
\Indm

\Return $\rho(S), \sigma(S)$\;
\end{algorithm}

Algorithm~\ref{alg:risk} formalizes the assignment of safety metadata to each scenario. The input $S=(H,E,M)$ decomposes a scenario into three blocks: hazard events $H$, environmental conditions $E$, and mission parameters $M$. The function $F(S)$ extracts the set of active hazards, such as icing events, GNSS jamming, or battery failures. If hazards are present, their maximum severity is recorded as $\rho_{\text{hazards}}(S)$; otherwise, this value defaults to~0. Environmental features are extracted as wind speed $v_{\text{wind}}$, visibility class $\gamma_{\text{vis}}$, and swarm size $n_{\text{swarm}}$, all of which are compared against operational thresholds $v_{\text{th}}, \gamma_{\text{th}}, n_{\text{th}}$ to compute an environmental risk contribution $\rho_{\text{env}}(S)$. The final risk score $\rho(S)$ is the maximum of hazard and environment contributions, thereby prioritizing catastrophic hazards while still capturing adverse operating conditions.

The second output, $\sigma(S)$, provides an interpretable categorical safety tag. The set $\Sigma$ is partitioned into domains such as \texttt{Weather} (e.g., icing, lightning), \texttt{Navigation} (e.g., GNSS spoofing, link loss), \texttt{Energy} (e.g., low battery, fuel exhaustion), and \texttt{Collision-Avoidance} (e.g., separation breaches in swarms). The assignment rule maps each scenario to the dominant category associated with its highest-severity hazard or environmental stressor. This two-level labeling framework produces not only a scalar risk level $\rho(S) \in \{0,1,2,3\}$ but also a categorical tag $\sigma(S) \in \Sigma$, enabling both coarse-grained benchmarking and fine-grained analysis of failure modes.

In summary, the risk and safety labeling step transforms raw scenario metadata into standardized, reproducible indicators. The quantitative score $\rho(S)$ facilitates statistical benchmarking across large datasets, while the categorical tag $\sigma(S)$ enhances interpretability by linking risk to root causes. Together, these labels make \texttt{UAVBench} suitable for evaluating agentic AI systems under both nominal and safety-critical conditions.

\begin{figure*}[h!]
  \centering
  \includegraphics[width=0.95\textwidth]{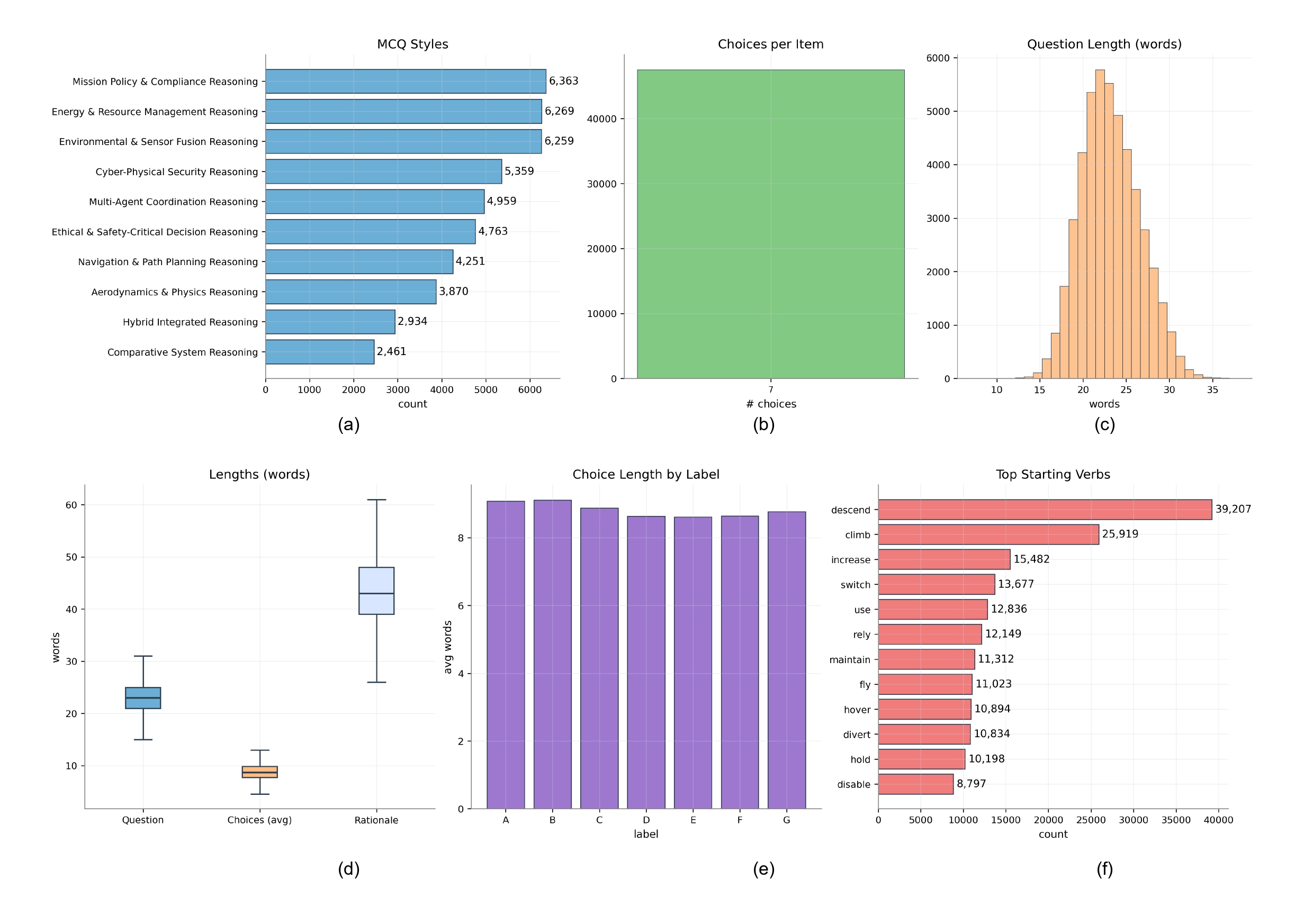}
  \caption{Overview of \texttt{UAVBench\_MCQ} dataset structure and linguistic statistics.
  (a) Distribution of multiple-choice question (MCQ) styles across reasoning domains;
  (b) Number of answer choices per question;
  (c) Distribution of question lengths in words;
  (d) Comparison of word counts for questions, averaged choices, and rationales;
  (e) Average choice length by option label (A–G); and
  (f) The most frequent starting verbs in the choice text.
  Together, these subfigures summarize the content balance, linguistic complexity, and stylistic diversity of \texttt{UAVBench\_MCQ} items.
  }
  \label{fig:uavbench_mcq_overview}
\end{figure*}

\begin{figure*}[htbp]
    \centering
    \includegraphics[width=1\textwidth]{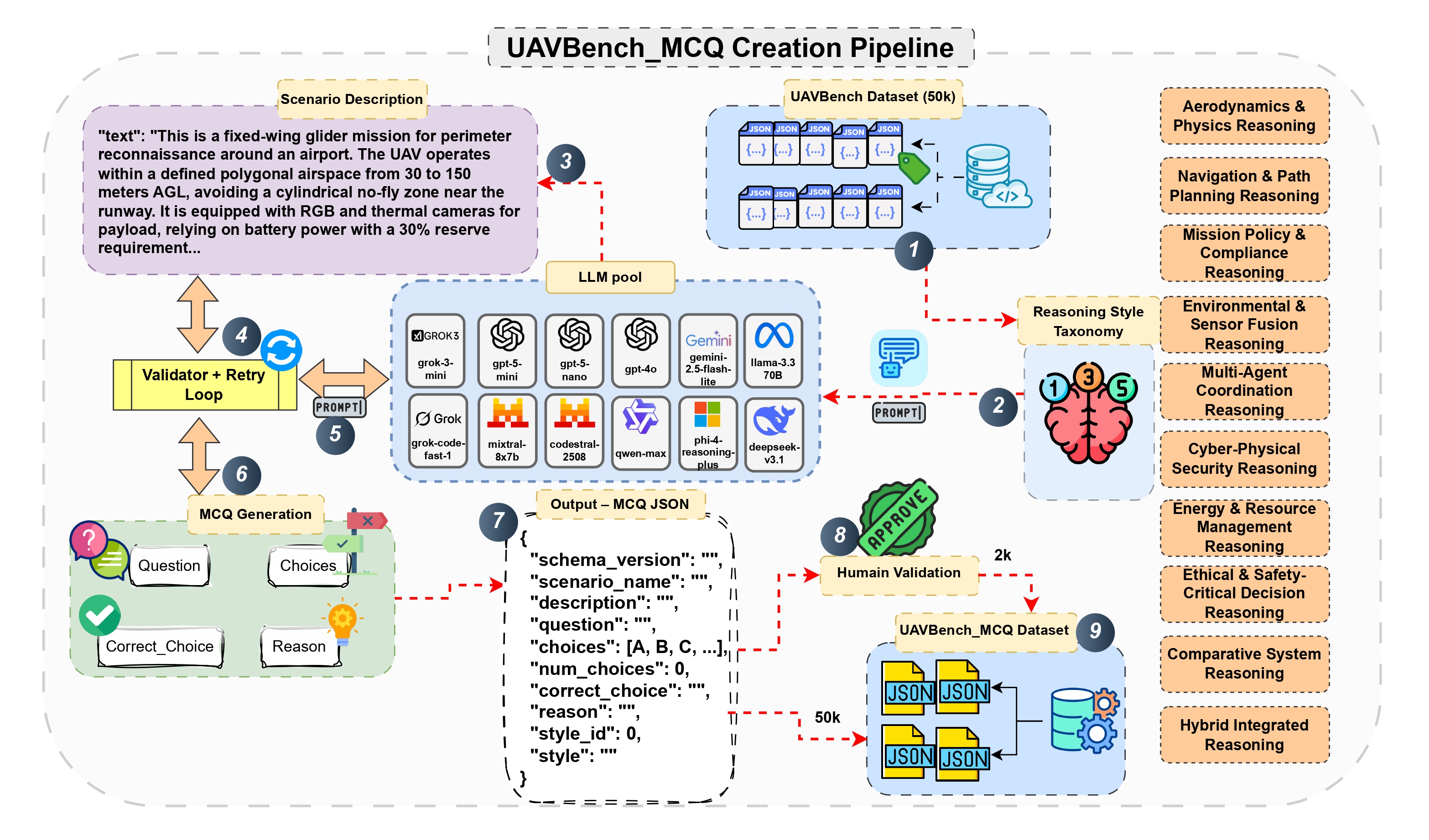} 
    \caption{\texttt{UAVBench\_MCQ} Creation Pipeline.}
    \label{fig:uavbench_mcq_pipeline}
\end{figure*}

\section{Experiments and Results}
\label{sec:exp_res}

This section presents the experimental evaluation of reasoning in agentic AI–driven UAV systems using \texttt{UAVBench} and its structured extension \texttt{UAVBench\_MCQ}. We outline the reasoning framework, describe the generation of structured MCQs, and report model performance across diverse reasoning domains to highlight current capabilities and remaining challenges in UAV autonomy.

\subsection{Reasoning Styles and \texttt{UAVBench\_MCQ} Framework}
\label{subsec:reasoning_framework}

To evaluate reasoning in agentic AI–driven UAV systems, we extend the dataset with \texttt{UAVBench\_MCQ}, a unified framework for structured reasoning and benchmarking. We define ten reasoning styles—covering aerodynamics, navigation, policy compliance, environmental sensing, multi-agent coordination, cyber-physical security, energy management, ethics and safety, comparative systems, and hybrid integration—each guided by a style-specific prompt and validation rule set. Validated \texttt{UAVBench} scenarios are transformed into self-contained JSON multiple-choice questions containing the scenario description, question, options, correct answer, rationale, and metadata. The framework enforces grounded realism, structural completeness, style-dependent option counts, and length limits to ensure consistent, large-scale, and programmatically gradable evaluation of UAV reasoning. Fig.~\ref{fig:uavbench_mcq_pipeline} illustrates this pipeline, in which each scenario is mapped to a reasoning style, processed with LLM-based prompts, validated for schema and logic, and stored as standardized JSON objects for reproducible benchmarking.

\begin{table*}[h!]
\centering
\caption{Reasoning styles defined in \texttt{UAVBench} for comprehensive evaluation of UAV cognitive and ethical reasoning.}
\label{tab:reasoning_styles}
\renewcommand{\arraystretch}{0.9}
\rowcolors{2}{white}{cyanblue!70}
\begin{tabular}{p{0.3cm} p{5.5cm} p{10cm}}
\toprule
\textbf{ID} & \textbf{Reasoning Style} & \textbf{Focus and Evaluation Scope} \\
\midrule
1 & Aerodynamics \& Physics Reasoning & Models flight mechanics, including lift, drag, thrust, and control stability. Evaluates physically plausible flight and awareness of aerodynamic constraints. \\
2 & Navigation \& Path Planning Reasoning & Tests trajectory optimization, obstacle avoidance, and spatial reasoning under time and energy constraints. \\
3 & Mission Policy \& Compliance Reasoning & Evaluates adherence to airspace regulations, operational limits, and mission rules (e.g., NFZ, BVLOS, privacy). \\
4 & Environmental \& Sensor Fusion Reasoning & Assesses understanding of environmental conditions, sensor fusion, and perception reliability under uncertainty. \\
5 & Multi-Agent Coordination Reasoning & Focuses on cooperative UAV behavior, communication, and deconfliction among multiple agents in dynamic environments. \\
6 & Cyber-Physical Security Reasoning & Evaluates response to spoofing, jamming, or sensor compromise, testing situational awareness and integrity preservation. \\
7 & Energy \& Resource Management Reasoning & Analyzes energy-efficient decision-making, load balancing, and mission prioritization under resource limitations. \\
8 & Ethical \& Safety-Critical Decision Reasoning & Captures moral trade-offs, safety-of-life priorities, lawful conduct, and responsible autonomy during emergencies. \\
9 & Comparative System Reasoning & Compares UAV designs, control strategies, or architectures to infer performance trade-offs and optimal configurations. \\
10 & Hybrid Integrated Reasoning & Integrates multiple reasoning domains (e.g., navigation + ethics + resource) to test multi-objective mission optimization. \\
\bottomrule
\end{tabular}
\end{table*}

\subsection{UAVBench\_MCQ: Structured Multi-Style MCQ Generation}
\label{subsec:UAVBench_mcq}

UAVBench\_MCQ transforms validated \texttt{UAVBench} scenarios into structured, interpretable, and machine-readable MCQs. Each MCQ is a self-contained JSON object that includes the scenario description, question, labeled options, the correct choice, rationale, and metadata such as style identifier, generator model, and schema version. The dataset is thus both human-interpretable and programmatically gradable, enabling large-scale benchmarking of UAV reasoning agents.

\begin{table*}[h!]
\centering
\caption{Notation used in the \texttt{UAVBench\_MCQ} generation process.}
\label{tab:notation_mcq}
\renewcommand{\arraystretch}{0.8}
\rowcolors{2}{white}{cyanblue!70}
\begin{tabular}{p{2.5cm} p{1.2cm} p{9.5cm}}
\toprule
\textbf{Symbol} & \textbf{Block} & \textbf{Definition} \\
\midrule
$\mathcal{S}$ & Input & Validated \texttt{UAVBench} scenario in structured JSON format. \\
$D$ & Input & Natural-language description derived from $\mathcal{S}$, summarizing mission type, UAV configuration, and constraints. \\
$S$ & Context & Reasoning style identifier $(1\!-\!10)$ guiding prompt selection and validation logic. \\
$Q$ & Output & MCQ question targeting reasoning consistent with style $S$. \\
$C = \{C_A, \ldots, C_G\}$ & Output & Labeled set of candidate options; typically four or seven depending on style. \\
$C^*$ & Output & Correct option satisfying physical, logical, or ethical constraints. \\
$R$ & Reasoning & Explanation or rationale justifying the correct choice. \\
$\rho(Q)$ & Risk & Embedded risk or severity level in the question context (low–critical). \\
$\Pi(D,S)$ & Mapping & Prompting function transforming $(D,S)$ into structured MCQ output. \\
$h$ & Metadata & Content hash for versioning and deduplication. \\
\bottomrule
\end{tabular}
\end{table*}

\subsubsection{MCQ Representation}
Each generated MCQ follows a standardized JSON schema:
\begin{equation}
q = \langle   D, S, Q, \mathcal{O}, i^*, R, h \rangle,
\end{equation}
where $\mathcal{O}$ represents the ordered set of candidate options, and $i^*$ the single correct answer. The schema ensures backward compatibility across updates, traceability to source scenarios, and interoperability with automated evaluation pipelines.

\subsubsection{Design Constraints}
To maintain reliability and interpretability across reasoning styles, the generation process enforces strict constraints:
\begin{itemize}
    \item \textit{Grounded realism:} Each MCQ must reference only facts available in the original scenario JSON.
    \item \textit{Structured completeness:} The fields \texttt{question}, \texttt{choices}, \texttt{correct\_choice}, and \texttt{reason} are mandatory.
    \item \textit{Consistency rules:} Exactly one correct option must exist; distractors must remain locally plausible but violate at least one constraint relevant to style $S$.
    \item \textit{Compactness:} Question length $\leq 28$ words; choice length $\leq 14$ words.
    \item \textit{Ethical schema:} For example, Style~8, seven options (A–G) are used to encode ethical trade-offs with explicit prioritization of human safety.
\end{itemize}

\begin{algorithm}[h!]
\caption{UAVBench\_MCQ Multi-Style Generation Pipeline}
\label{alg:UAVBench_mcq}
\KwIn{Validated scenario $\mathcal{S}$, reasoning style $S$}
\KwOut{Structured MCQ object $q$ linked to $\mathcal{S}$}

\tcp{Stage 1: Scenario Description}
$D \leftarrow$ \textsc{GenerateDescription}$(\mathcal{S})$\;
\Indp
Invoke LLM with a style-specific system prompt to produce a concise description ( $\leq$10 sentences).\;
Sanitize and attach description to $\mathcal{S}$ for contextual grounding.\;
\Indm

\tcp{Stage 2: MCQ Generation}
$q' \leftarrow$ \textsc{MakeMCQ}$(D,S)$ using style-specific prompt template.\;
\Indp
Extract fields $\{Q, \mathcal{O}, i^*, R\}$; validate schema, label format, and distinct options.\;
If validation fails, retry up to $R=3$ iterations.\;
\Indm

\tcp{Stage 3: Metadata and Persistence}
Compute hash $h$ and assemble $q = \langle   D, S, Q, \mathcal{O}, i^*, R, h \rangle$.\;
Save as \texttt{\textless scenario\_name\textgreater\_\{S\}\_\{h\}\_mcq.json}.\;
\Return $q$.
\end{algorithm}

Algorithm~\ref{alg:UAVBench_mcq} illustrates a modular, style-driven generation pipeline that enforces format, validity, and realism. The design separates descriptive grounding from question synthesis, ensuring that each LLM instance focuses on reasoning rather than scenario rewriting. The retry mechanism and schema validation safeguard against malformed or logically inconsistent outputs, while the hash-based persistence guarantees reproducibility and deduplication across large-scale generations.

UAVBench\_MCQ provides an interpretable and standardized bridge between simulation-grounded UAV data and reasoning-based evaluation. By incorporating ten reasoning styles and structured JSON outputs, it enables both quantitative benchmarking and qualitative insight into LLMs’ decision integrity. The schema’s inclusion of metadata (e.g., \texttt{schema\_version}, \texttt{style\_id}, and \texttt{hash}) ensures transparent provenance tracking and long-term dataset evolution. Ultimately, \texttt{UAVBench\_MCQ} advances the evaluation of UAV autonomy by combining physical realism, cognitive depth, and ethical accountability within a reproducible benchmarking ecosystem.

\begin{table*}[h!]
\centering
\caption{Accuracy (\%) on \textit{Perception \& Physical World} reasoning styles in UAVBench.}
\label{tab:perception_physical}
\renewcommand{\arraystretch}{0.9}
\rowcolors{2}{white}{cyanblue!70}
\scriptsize
\begin{tabular}{l l l l
S[table-format=3.1]
S[table-format=3.1]
S[table-format=3.1]}
\toprule
\rowcolor{gray!35}
\textbf{Model} & \textbf{Company} & \textbf{Size} & \textbf{License} &
\textbf{(1) Aerodynamics \& Physics} & \textbf{(4) Environmental \& Sensor Fusion} & \textbf{Avg.} \\
\midrule
Qwen3 235B A22B (2507) & Alibaba & 235B & Open & 82.5 & 97.0 & 89.8 \\
ChatGPT 4o & OpenAI & N/A & Proprietary & 74.5 & 96.5 & 85.5 \\
GPT-5 Chat & OpenAI & N/A & Proprietary & 73.5 & 97.0 & 85.3 \\
Qwen3 Max & Alibaba & N/A & Open & 73.5 & 96.0 & 84.8 \\
Mistral Medium 3.1 & Mistral AI & N/A & Proprietary & 72.5 & 94.5 & 83.5 \\
ERNIE 4.5 300B A47B & Baidu & 300B & Open & 71.0 & 96.0 & 83.5 \\
GPT-4.1 Mini & OpenAI & N/A & Proprietary & 68.0 & 97.5 & 82.8 \\
InternVL3 78B & OpenGVLab & 78B & Open & 69.0 & 96.5 & 82.8 \\
GPT-4.1 & OpenAI & N/A & Proprietary & 69.5 & 95.5 & 82.5 \\
GPT-4.1 & OpenAI & N/A & Proprietary & 69.5 & 95.5 & 82.5 \\
Kimi K2 & Moonshot AI & 1T & Open & 69.5 & 95.0 & 82.3 \\
Claude-haiku-4.5 & Anthropic & N/A & Proprietary & 68.0 & 94.5 & 81.3 \\
Phi 4 Reasoning Plus & Microsoft & 14B & Open & 65.5 & 97.0 & 81.3 \\
Gemini 2.5 Flash & Google & 391B & Proprietary & 65.5 & 96.0 & 80.8 \\
Qwen3 VL 8B Instruct & Alibaba & 8B & Open & 64.5 & 96.5 & 80.5 \\
DeepSeek Chat V3 (0324) & DeepSeek & 685B & Open & 65.0 & 95.5 & 80.3 \\
DeepSeek V3.1 Terminus & DeepSeek & N/A & Open & 62.5 & 94.5 & 78.5 \\
DeepSeek V3.2 Exp & DeepSeek & N/A & Open & 61.0 & 95.0 & 78.0 \\
Llama-4-scout & Meta & 17B & Open & 59.0 & 96.5 & 77.8 \\
Grok 4 Fast & xAI & N/A & Proprietary & 60.0 & 89.5 & 74.8 \\
Qwen 2.5 7B Instruct & Alibaba & 7B & Open & 54.5 & 91.0 & 72.8 \\
LFM 2 2.6B & Liquid AI & 2.6B & Open & 49.0 & 95.5 & 72.3 \\
Gemma-3n-e4b-it & Google & 4B & Open & 49.0 & 94.5 & 71.8 \\
Olmo 2 32B Instruct & AllenAI & 32B & Open & 49.0 & 91.5 & 70.3 \\
LFM2-8B-A1B & Liquid AI & 8B & Open & 47.0 & 92.0 & 69.5 \\
Llama 3.1 8B Instruct & Meta & 8B & Open & 46.0 & 92.5 & 69.3 \\
Jamba-mini-1.7 & AI21 & N/A & Open & 48.5 & 86.0 & 67.3 \\
Llama 3.2 3B Instruct & Meta & 3B & Open & 37.5 & 86.5 & 62.0 \\
Granite-4.0-h-micro & IBM & 3B & Open & 36.0 & 87.0 & 61.5 \\
Claude Sonnet 4.5 & Anthropic & 468B & Proprietary & 36.0 & 81.0 & 58.5 \\
GLM-4.6 & Z.AI & 357B & Open & 34.5 & 36.5 & 35.5 \\
Qwen3-30B-A3B & Alibaba & 30B & Open & 3.5 & 7.5 & 5.5 \\
Nemotron Nano 9B V2 & NVIDIA & 9B & Open & 3.0 & 0.0 & 1.5 \\
\bottomrule
\end{tabular}\\
\noindent\textit{LLM parameters:} \texttt{top\_p = 1.0}, which is the nucleus sampling parameter (1.0 = all tokens considered); \texttt{max\_tokens = 16}, which defines the maximum number of tokens generated; \texttt{temperature = 0.0}, which controls randomness (0 = deterministic output); and \texttt{max\_retries = 5}, which specifies the maximum number of retry attempts in case of LLM failure.
\end{table*}

Fig.~\ref{fig:uavbench_mcq_overview} shows the composition and linguistic patterns of the \texttt{UAVBench\_MCQ} dataset.
It includes distributions of question styles, number of answer choices, and word-length statistics for questions, choices, and rationales.
The overall balance and variety across reasoning styles, lexical structures, and choice formulations illustrate the dataset’s breadth and quality for UAV-related reasoning tasks.

\subsection{Evaluation Metrics}
\label{subsec:evaluation_metrics}

To comprehensively assess reasoning performance across diverse UAV mission domains, we adopt a four-metric evaluation framework that captures both overall correctness and cross-style consistency. While raw \textit{Accuracy} measures general task performance, it fails to distinguish between models that perform well in certain reasoning styles but poorly in others. To address this, we introduce three complementary statistics—\textit{Mean Accuracy}, \textit{Standard Deviation}, and the \textit{Balanced Style Score (BSS)}—computed from the per-style accuracies $\{a_s\}_{s=1}^{S}$, where $S=10$ denotes the number of reasoning styles defined in \texttt{UAVBench\_MCQ}.

\paragraph{Accuracy (\%)}
Overall accuracy measures the proportion of correctly answered multiple-choice questions across the entire benchmark:
\begin{equation}
\mathrm{Accuracy} = \frac{N_{\text{correct}}}{N_{\text{total}}} \times 100,
\end{equation}
where $N_{\text{correct}}$ and $N_{\text{total}}$ denote the number of correct responses and total questions, respectively. This metric captures aggregate correctness independent of reasoning style.

\paragraph{Mean Accuracy (\%)}
To evaluate average performance across reasoning categories, we compute the mean of per-style accuracies:
\begin{equation}
\overline{a} = \frac{1}{S}\sum_{s=1}^{S} a_s,
\end{equation}
where $a_s$ represents the accuracy for reasoning style $s$. High $\overline{a}$ values indicate generally strong performance across all cognitive dimensions of UAV reasoning.

\paragraph{Standard Deviation (\%)}
To quantify performance consistency, we calculate the standard deviation of accuracies across all reasoning styles:
\begin{equation}
\sigma(a) = \sqrt{\frac{1}{S}\sum_{s=1}^{S}(a_s - \overline{a})^2}.
\end{equation}
Lower $\sigma(a)$ values indicate balanced reasoning ability, while higher values reveal specialization or weakness in certain domains (e.g., strong in physics but weak in ethics).

\paragraph{Balanced Style Score (BSS)}
Finally, we propose the \textit{Balanced Style Score} (BSS) as an integrated indicator of both accuracy and balance. BSS combines the geometric mean of per-style accuracies with a penalty term for imbalance:
\begin{equation}
\mathrm{BSS} = 
\left(\prod_{s=1}^{S}(a_s + \varepsilon)^{1/S}\right)
\times 
\left(1 - \frac{\sigma(a)}{\overline{a}}\right),
\end{equation}
where $\varepsilon$ is a small constant ($10^{-6}$) to avoid undefined logarithms. The first term rewards uniformly high performance, while the second penalizes uneven distribution across reasoning styles. BSS values lie in $[0,1]$, with higher scores indicating both accurate and consistent reasoning—a key property for safe, reliable UAV autonomy.

Together, these four metrics provide a multidimensional assessment of LLM reasoning under UAVBench\_MCQ. \textit{Accuracy} captures raw task success, \textit{Mean Accuracy} reflects general competence, \textit{Standard Deviation} measures balance across reasoning styles, and \textit{BSS} synthesizes them into a single interpretable metric that rewards models exhibiting both correctness and cross-domain consistency—an essential criterion for trustworthy UAV decision-making.

\subsection{Performance on Perception and Physical World Reasoning}

Table~\ref{tab:perception_physical} presents the results of model performance on UAVBench’s \textit{Perception \& Physical World} reasoning tasks, which assess a model’s capability to understand aerodynamics, environmental dynamics, and sensor fusion scenarios. Among all evaluated systems, \textit{Qwen3 235B A22B} achieves the highest average accuracy of \textit{89.8\%}, outperforming leading proprietary models such as \textit{ChatGPT 4o} (85.5\%) and \textit{GPT-5 Chat} (85.3\%). Open-source models from Alibaba, including Qwen3 Max and Qwen3 VL 8B Instruct, consistently rank among the top performers, indicating the growing competitiveness of open models in physics-grounded reasoning. Proprietary systems from OpenAI and Mistral also demonstrate strong and stable results across both reasoning categories, suggesting robust internalization of physical and environmental relationships even in UAV-specific contexts.

Smaller, lightweight open models (e.g., \textit{Llama 3.1 8B}, \textit{Gemma-3n-e4b-it}, and \textit{Olmo 2 32B}) exhibit a marked decline in performance, with average accuracies ranging from 61--70\%. This trend suggests that reasoning over aerial dynamics and sensor-based perception remains highly dependent on model scale and domain-specific training. At the lower end, \textit{Nemotron Nano 9B V2} and \textit{Qwen3-30B-A3B} perform poorly (below 6\%), revealing limited generalization to grounded physical reasoning. Across nearly all models, accuracies on \textit{Environmental \& Sensor Fusion} tasks exceed those on \textit{Aerodynamics \& Physics}, implying that current LLMs integrate perceptual and multimodal cues more effectively than they infer dynamic physical laws. Overall, these findings indicate that while large-scale, instruction-tuned models—both open and proprietary—are achieving near-human reliability in perceptual reasoning, mastering fine-grained aerodynamics and UAV physics remains an open research challenge.

\begin{table*}[h!]
\centering
\caption{Accuracy (\%) on \textit{Planning, Coordination \& Resources} reasoning styles in UAVBench.}
\label{tab:planning_coordination}
\renewcommand{\arraystretch}{0.9}
\rowcolors{2}{white}{cyanblue!70}
\scriptsize
\begin{tabular}{l l l l
S[table-format=3.1]
S[table-format=3.1]
S[table-format=3.1]
S[table-format=3.1]}
\toprule
\rowcolor{gray!35}
\textbf{Model} & \textbf{Company} & \textbf{Size} & \textbf{License} &
\textbf{(2) Navigation \& Path} & \textbf{(5) Multi-Agent Coord.} & \textbf{(7) Energy \& Resource} & \textbf{Avg.} \\
\midrule
Qwen3 235B A22B (2507) & Alibaba & 235B & Open & 81.5 & 76.5 & 71.5 & 76.5 \\
GPT-5 Chat & OpenAI & N/A & Proprietary & 78.0 & 72.0 & 68.5 & 72.8 \\
ChatGPT 4o & OpenAI & N/A & Proprietary & 80.5 & 70.0 & 64.5 & 71.7 \\
Qwen3 Max & Alibaba & N/A & Open & 77.0 & 70.5 & 65.0 & 70.8 \\
GPT-4.1 & OpenAI & N/A & Proprietary & 82.5 & 67.0 & 62.5 & 70.7 \\
GPT-4.1 Mini & OpenAI & N/A & Proprietary & 75.5 & 71.0 & 64.5 & 70.3 \\
Phi 4 Reasoning Plus & Microsoft & 14B & Open & 76.5 & 67.0 & 67.0 & 70.2 \\
Kimi K2 & Moonshot AI & 1T & Open & 67.5 & 71.5 & 70.0 & 69.7 \\
Gemini 2.5 Flash & Google & 391B & Proprietary & 73.5 & 69.5 & 66.0 & 69.7 \\
InternVL3 78B & OpenGVLab & 78B & Open & 71.5 & 67.5 & 68.0 & 69.0 \\
Llama-4-scout & Meta & 17B & Open & 72.0 & 71.5 & 63.5 & 69.0 \\
Qwen3 VL 8B Instruct & Alibaba & 8B & Open & 76.0 & 64.5 & 66.0 & 68.8 \\
Claude-haiku-4.5 & Anthropic & N/A & Proprietary & 73.0 & 68.0 & 62.5 & 67.8 \\
ERNIE 4.5 300B A47B & Baidu & 300B & Open & 71.5 & 68.0 & 63.0 & 67.5 \\
Mistral Medium 3.1 & Mistral AI & N/A & Proprietary & 69.0 & 67.5 & 65.0 & 67.2 \\
Gemma-3n-e4b-it & Google & 4B & Open & 63.5 & 63.5 & 71.0 & 66.0 \\
DeepSeek Chat V3 (0324) & DeepSeek & 685B & Open & 68.5 & 65.5 & 63.0 & 65.7 \\
Grok 4 Fast & xAI & N/A & Proprietary & 69.5 & 59.5 & 58.0 & 62.3 \\
DeepSeek V3.2 Exp & DeepSeek & N/A & Open & 63.0 & 62.5 & 61.0 & 62.2 \\
LFM 2 2.6B & Liquid AI & 2.6B & Open & 65.0 & 62.0 & 55.5 & 60.8 \\
DeepSeek V3.1 Terminus & DeepSeek & N/A & Open & 59.5 & 58.0 & 62.5 & 60.0 \\
Llama 3.1 8B Instruct & Meta & 8B & Open & 57.0 & 58.5 & 59.5 & 58.3 \\
Qwen 2.5 7B Instruct & Alibaba & 7B & Open & 60.5 & 52.0 & 61.0 & 57.8 \\
Llama 3.2 3B Instruct & Meta & 3B & Open & 54.0 & 55.0 & 59.0 & 56.0 \\
Olmo 2 32B Instruct & AllenAI & 32B & Open & 57.0 & 60.0 & 51.0 & 56.0 \\
LFM2-8B-A1B & Liquid AI & 8B & Open & 65.5 & 57.0 & 41.0 & 54.5 \\
Jamba-mini-1.7 & AI21 & N/A & Open & 54.5 & 55.0 & 43.5 & 51.0 \\
Claude Sonnet 4.5 & Anthropic & 468B & Proprietary & 53.0 & 50.5 & 41.0 & 48.2 \\
Granite-4.0-h-micro & IBM & 3B & Open & 50.0 & 43.5 & 50.5 & 48.0 \\
GLM-4.6 & Z.AI & 357B & Open & 31.5 & 47.5 & 32.0 & 37.0 \\
Qwen3-30B-A3B & Alibaba & 30B & Open & 4.5 & 6.5 & 4.0 & 5.0 \\
Nemotron Nano 9B V2 & NVIDIA & 9B & Open & 1.0 & 0.5 & 1.0 & 0.8 \\
\bottomrule
\end{tabular}\\
\noindent\textit{LLM parameters:} \texttt{top\_p = 1.0}, which is the nucleus sampling parameter (1.0 = all tokens considered); \texttt{max\_tokens = 16}, which defines the maximum number of tokens generated; \texttt{temperature = 0.0}, which controls randomness (0 = deterministic output); and \texttt{max\_retries = 5}, which specifies the maximum number of retry attempts in case of LLM failure.
\end{table*}

\subsection{Performance on Planning, Coordination, and Resource Reasoning}

Table~\ref{tab:planning_coordination} reports the accuracy of leading LLMs on UAVBench’s \textit{Planning, Coordination \& Resources} reasoning tasks, encompassing \textit{Navigation \& Path Planning}, \textit{Multi-Agent Coordination}, and \textit{Energy \& Resource Management}. The results indicate that \textit{Qwen3 235B A22B} again achieves the highest overall performance with an average accuracy of \textit{76.5\%}, demonstrating balanced competence across trajectory optimization, obstacle avoidance, and energy-aware planning. Proprietary models such as \textit{GPT-5 Chat} (72.8\%) and \textit{ChatGPT 4o} (71.7\%) follow closely, reflecting their strength in dynamic decision-making and temporal-spatial reasoning. Open-source systems like \textit{Qwen3 Max} and \textit{Phi 4 Reasoning Plus} also perform competitively, suggesting that well-tuned open models are closing the gap in complex reasoning domains. In contrast, \textit{GPT-4.1} exhibits notably strong navigation performance (82.5\%) but comparatively weaker coordination and resource management, suggesting a bias toward single-agent spatial reasoning.

Performance trends across subtasks reveal that \textit{Navigation \& Path Planning} generally yields higher accuracies than the other two categories, emphasizing that most LLMs handle structured spatial reasoning better than cooperative or resource-constrained scenarios. \textit{Multi-Agent Coordination} and \textit{Energy \& Resource Management} tasks, which require distributed decision-making and trade-off optimization, remain challenging across all models, with even top performers achieving below 80\%. Smaller open models such as \textit{Llama 3.1 8B}, \textit{Gemma-3n-e4b-it}, and \textit{DeepSeek V3.2 Exp} average between 56–66\%, while lightweight architectures like \textit{Nemotron Nano 9B V2} and \textit{Qwen3-30B-A3B} fall below 6\%. These results collectively suggest that while frontier models demonstrate emerging capabilities in autonomous planning, true competence in cooperative multi-agent coordination and energy-aware mission optimization remains an open research frontier for both open and proprietary LLMs.

\begin{table*}[h!]
\centering
\caption{Accuracy (\%) on \textit{Governance, Ethics \& Security} reasoning styles in UAVBench.}
\label{tab:governance_ethics}
\renewcommand{\arraystretch}{0.9}
\rowcolors{2}{white}{cyanblue!70}
\scriptsize
\begin{tabular}{l l l l
S[table-format=3.1]
S[table-format=3.1]
S[table-format=3.1]
S[table-format=3.1]}
\toprule
\rowcolor{gray!35}
\textbf{Model} & \textbf{Company} & \textbf{Size} & \textbf{License} &
\textbf{(3) Policy \& Compliance} & \textbf{(8) Ethical \& Safety-Critical} & \textbf{(6) Cyber-Physical Sec.} & \textbf{Avg.} \\
\midrule
Qwen3 235B A22B (2507) & Alibaba & 235B & Open & 76.0 & 75.5 & 96.5 & 82.7 \\
ChatGPT 4o & OpenAI & N/A & Proprietary & 72.0 & 73.0 & 97.0 & 80.7 \\
Qwen3 Max & Alibaba & N/A & Open & 68.5 & 76.0 & 97.0 & 80.5 \\
GPT-5 Chat & OpenAI & N/A & Proprietary & 65.5 & 76.0 & 98.0 & 79.8 \\
GPT-4.1 & OpenAI & N/A & Proprietary & 73.0 & 70.0 & 96.0 & 79.7 \\
DeepSeek Chat V3 (0324) & DeepSeek & 685B & Open & 66.0 & 75.5 & 95.0 & 78.8 \\
DeepSeek V3.2 Exp & DeepSeek & N/A & Open & 61.0 & 77.5 & 96.0 & 78.2 \\
Kimi K2 & Moonshot AI & 1T & Open & 69.0 & 68.5 & 96.5 & 78.0 \\
GPT-4.1 Mini & OpenAI & N/A & Proprietary & 68.0 & 67.0 & 97.5 & 77.5 \\
Gemini 2.5 Flash & Google & 391B & Proprietary & 62.0 & 71.5 & 97.0 & 76.8 \\
InternVL3 78B & OpenGVLab & 78B & Open & 62.5 & 72.0 & 96.0 & 76.8 \\
Mistral Medium 3.1 & Mistral AI & N/A & Proprietary & 59.0 & 75.0 & 96.5 & 76.8 \\
Claude-haiku-4.5 & Anthropic & N/A & Proprietary & 68.0 & 67.0 & 95.0 & 76.7 \\
DeepSeek V3.1 Terminus & DeepSeek & N/A & Open & 59.5 & 72.0 & 96.0 & 75.8 \\
Gemma-3n-e4b-it & Google & 4B & Open & 61.5 & 72.0 & 95.0 & 76.2 \\
Phi 4 Reasoning Plus & Microsoft & 14B & Open & 57.0 & 73.5 & 96.5 & 75.7 \\
Grok 4 Fast & xAI & N/A & Proprietary & 60.5 & 69.5 & 94.0 & 74.7 \\
Llama-4-scout & Meta & 17B & Open & 63.0 & 63.5 & 96.0 & 74.2 \\
ERNIE 4.5 300B A47B & Baidu & 300B & Open & 59.5 & 68.0 & 94.5 & 74.0 \\
Qwen3 VL 8B Instruct & Alibaba & 8B & Open & 62.5 & 62.0 & 97.0 & 73.8 \\
LFM 2 2.6B & Liquid AI & 2.6B & Open & 57.5 & 57.5 & 94.5 & 69.8 \\
Olmo 2 32B Instruct & AllenAI & 32B & Open & 54.5 & 61.5 & 88.5 & 68.2 \\
LFM2-8B-A1B & Liquid AI & 8B & Open & 47.0 & 57.5 & 95.0 & 66.5 \\
Qwen 2.5 7B Instruct & Alibaba & 7B & Open & 47.0 & 53.5 & 93.0 & 64.5 \\
Llama 3.1 8B Instruct & Meta & 8B & Open & 45.0 & 57.0 & 91.0 & 64.3 \\
Claude Sonnet 4.5 & Anthropic & 468B & Proprietary & 48.0 & 49.0 & 91.0 & 62.7 \\
Llama 3.2 3B Instruct & Meta & 3B & Open & 43.5 & 47.5 & 87.5 & 59.5 \\
Granite-4.0-h-micro & IBM & 3B & Open & 37.5 & 54.0 & 87.0 & 59.5 \\
Jamba-mini-1.7 & AI21 & N/A & Open & 32.0 & 45.0 & 88.5 & 55.2 \\
GLM-4.6 & Z.AI & 357B & Open & 41.5 & 41.5 & 66.5 & 49.8 \\
Qwen3-30B-A3B & Alibaba & 30B & Open & 4.0 & 6.0 & 10.0 & 6.7 \\
Nemotron Nano 9B V2 & NVIDIA & 9B & Open & 1.5 & 2.5 & 11.0 & 5.0 \\
\bottomrule
\end{tabular}\\
\noindent\textit{LLM parameters:} \texttt{top\_p = 1.0}, which is the nucleus sampling parameter (1.0 = all tokens considered); \texttt{max\_tokens = 16}, which defines the maximum number of tokens generated; \texttt{temperature = 0.0}, which controls randomness (0 = deterministic output); and \texttt{max\_retries = 5}, which specifies the maximum number of retry attempts in case of LLM failure.

\end{table*}

\begin{table*}[h!]
\centering
\caption{Accuracy (\%) on \textit{Systems \& Integration} reasoning styles in UAVBench.}
\label{tab:systems_integration}
\renewcommand{\arraystretch}{0.9}
\rowcolors{2}{white}{cyanblue!70}
\scriptsize
\begin{tabular}{l l l l
S[table-format=3.1]
S[table-format=3.1]
S[table-format=3.1]}
\toprule
\rowcolor{gray!35}
\textbf{Model} & \textbf{Company} & \textbf{Size} & \textbf{License} &
\textbf{(9) Comparative System} & \textbf{(10) Hybrid Integrated} & \textbf{Avg.} \\
\midrule
Qwen3 235B A22B (2507) & Alibaba & 235B & Open & 95.5 & 83.0 & 89.3 \\
Qwen3 Max & Alibaba & N/A & Open & 96.5 & 78.5 & 87.5 \\
ChatGPT 4o & OpenAI & N/A & Proprietary & 96.5 & 79.0 & 87.8 \\
Claude-haiku-4.5 & Anthropic & N/A & Proprietary & 94.0 & 80.5 & 87.3 \\
GPT-4.1 & OpenAI & N/A & Proprietary & 97.0 & 77.5 & 87.3 \\
GPT-5 Chat & OpenAI & N/A & Proprietary & 95.5 & 77.5 & 86.5 \\
GPT-4.1 Mini & OpenAI & N/A & Proprietary & 95.0 & 77.0 & 86.0 \\
Qwen3 VL 8B Instruct & Alibaba & 8B & Open & 94.5 & 76.0 & 85.3 \\
Kimi K2 & Moonshot AI & 1T & Open & 96.0 & 74.0 & 85.0 \\
Mistral Medium 3.1 & Mistral AI & N/A & Proprietary & 95.0 & 74.5 & 84.8 \\
InternVL3 78B & OpenGVLab & 78B & Open & 94.0 & 74.0 & 84.0 \\
Phi 4 Reasoning Plus & Microsoft & 14B & Open & 93.5 & 74.0 & 83.8 \\
Gemini 2.5 Flash & Google & 391B & Proprietary & 91.5 & 75.0 & 83.3 \\
Llama-4-scout & Meta & 17B & Open & 92.5 & 73.5 & 83.0 \\
Grok 4 Fast & xAI & N/A & Proprietary & 96.0 & 69.5 & 82.8 \\
DeepSeek Chat V3 (0324) & DeepSeek & 685B & Open & 93.5 & 71.5 & 82.5 \\
ERNIE 4.5 300B A47B & Baidu & 300B & Open & 93.5 & 69.5 & 81.5 \\
DeepSeek V3.1 Terminus & DeepSeek & N/A & Open & 91.5 & 71.0 & 81.3 \\
Gemma-3n-e4b-it & Google & 4B & Open & 91.5 & 71.0 & 81.3 \\
LFM 2 2.6B & Liquid AI & 2.6B & Open & 89.0 & 72.0 & 80.5 \\
DeepSeek V3.2 Exp & DeepSeek & N/A & Open & 91.0 & 67.0 & 79.0 \\
LFM2-8B-A1B & Liquid AI & 8B & Open & 91.0 & 65.0 & 78.0 \\
Llama 3.2 3B Instruct & Meta & 3B & Open & 86.5 & 63.0 & 74.8 \\
Qwen 2.5 7B Instruct & Alibaba & 7B & Open & 88.5 & 59.5 & 74.0 \\
Llama 3.1 8B Instruct & Meta & 8B & Open & 90.0 & 56.5 & 73.3 \\
Claude Sonnet 4.5 & Anthropic & 468B & Proprietary & 79.5 & 55.0 & 67.3 \\
Granite-4.0-h-micro & IBM & 3B & Open & 86.0 & 46.5 & 66.3 \\
Olmo 2 32B Instruct & AllenAI & 32B & Open & 84.5 & 58.0 & 71.3 \\
Jamba-mini-1.7 & AI21 & N/A & Open & 85.5 & 54.5 & 70.0 \\
GLM-4.6 & Z.AI & 357B & Open & 48.0 & 37.5 & 42.8 \\
Qwen3-30B-A3B & Alibaba & 30B & Open & 3.0 & 5.5 & 4.3 \\
Nemotron Nano 9B V2 & NVIDIA & 9B & Open & 1.5 & 2.0 & 1.8 \\
\bottomrule
\end{tabular}\\
\noindent\textit{LLM parameters:} \texttt{top\_p = 1.0}, which is the nucleus sampling parameter (1.0 = all tokens considered); \texttt{max\_tokens = 16}, which defines the maximum number of tokens generated; \texttt{temperature = 0.0}, which controls randomness (0 = deterministic output); and \texttt{max\_retries = 5}, which specifies the maximum number of retry attempts in case of LLM failure.
\end{table*}

\begin{figure*}[h!]
  \centering
  \includegraphics[width=\textwidth]{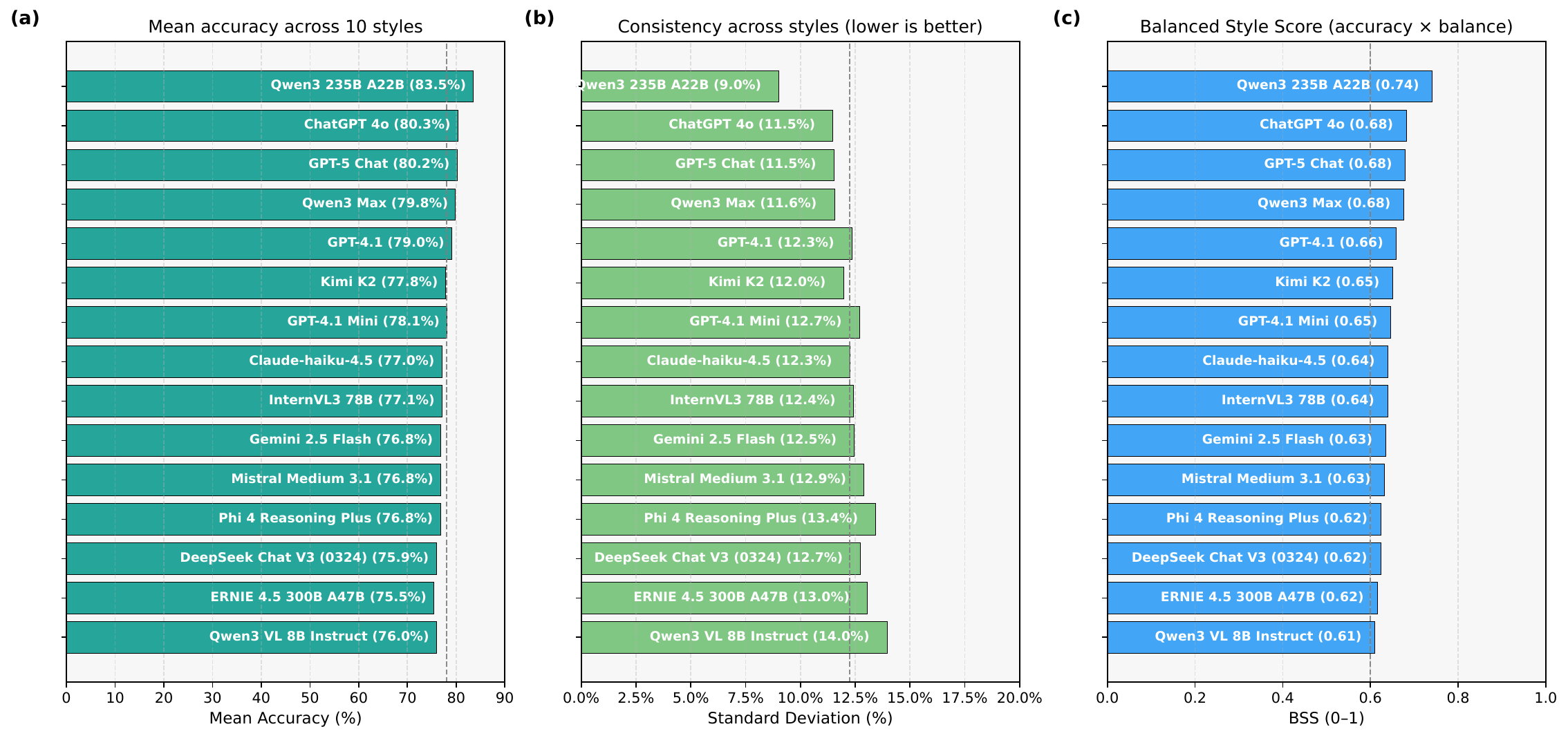}
  \caption{Top 15 UAVBench\_MCQ models ranked by Balanced Style Score (BSS).
  (a) Mean accuracy across ten reasoning styles, 
  (b) cross-style consistency measured as the standard deviation of accuracies (lower is better; axis in \%), 
  and (c) BSS integrating both accuracy and consistency.}
  \label{fig:uavbench_triptych_bss}
\end{figure*}

\subsection{Performance on Governance, Ethics, and Security Reasoning}

Table~\ref{tab:governance_ethics} summarizes model performance on UAVBench’s \textit{Governance, Ethics \& Security} reasoning tasks, which evaluate compliance with mission regulations, ethical decision-making under safety-critical conditions, and robustness against cyber-physical threats. The \textit{Qwen3 235B A22B} model leads with an average accuracy of \textit{82.7\%}, demonstrating exceptional competence in enforcing airspace policy and making high-stakes decisions. Proprietary models such as \textit{ChatGPT 4o} (80.7\%) and \textit{GPT-5 Chat} (79.8\%) closely follow, confirming their strength in ethical reasoning and operational rule interpretation. Open-source competitors like \textit{Qwen3 Max} (80.5\%) and \textit{DeepSeek Chat V3} (78.8\%) also perform robustly, demonstrating that governance-related reasoning is increasingly tractable for large open models. Interestingly, all high-performing models show particularly strong accuracy in \textit{Cyber-Physical Security} reasoning (95–98\%), suggesting that integrity-preservation and threat-response scenarios are well-captured in large-scale pretraining corpora.

Across subtasks, however, \textit{Mission Policy \& Compliance} and \textit{Ethical \& Safety-Critical Decision} reasoning remain more challenging than security-focused reasoning. Even top-tier models exhibit a noticeable performance gap—approximately 20 percentage points—between regulatory or moral judgment and cyber-physical threat handling. This indicates that while LLMs can recognize and describe technical countermeasures (e.g., against spoofing or jamming), they still struggle with normative constraints, lawful autonomy, and ethical trade-offs under uncertainty. Smaller open models (e.g., \textit{Llama 3.1 8B}, \textit{Qwen 2.5 7B}, and \textit{Olmo 2 32B}) yield averages between 60–65\%, reflecting their limited abstraction capacity for contextually nuanced or policy-dependent reasoning. Overall, the results suggest that while modern LLMs have made major strides in UAV security interpretation, achieving human-level ethical alignment and mission-compliance awareness remains a critical and unsolved dimension of safe autonomous operation.

\subsection{Performance on Systems and Integration Reasoning}

Table~\ref{tab:systems_integration} presents \texttt{UAVBench} results for \textit{Systems \& Integration} reasoning, which includes \textit{Comparative System Reasoning} and \textit{Hybrid Integrated Reasoning}. These categories assess a model’s capacity to compare UAV architectures, control designs, and mission configurations while optimizing across multiple reasoning domains. The top-performing model, \textit{Qwen3 235B A22B}, attains an impressive \textit{89.3\%} average accuracy, demonstrating a strong holistic understanding of UAV system trade-offs and integration principles. Close competitors such as \textit{ChatGPT 4o} (87.8\%), \textit{Qwen3 Max} (87.5\%), and \textit{GPT-4.1} (87.3\%) exhibit similarly high proficiency, indicating that both open and proprietary models have achieved mature competency in systems-level reasoning. Notably, \textit{Comparative System} reasoning yields the highest individual accuracies across all models—often exceeding 95\%—suggesting that performance evaluation and architecture comparison are well-aligned with the statistical and analytic strengths of large LLMs.

However, the more complex \textit{Hybrid Integrated Reasoning} task, which requires blending ethical, navigational, and resource-related reasoning to optimize multi-objective missions, remains a consistent bottleneck. Even the best-performing models score between 77–83\%, underscoring the difficulty of integrating heterogeneous reasoning modes into cohesive decisions. Mid-range open models such as \textit{Gemini 2.5 Flash}, \textit{InternVL3 78B}, and \textit{Phi 4 Reasoning Plus} maintain averages around 83–84\%, while smaller architectures like \textit{Llama 3.1 8B} and \textit{Qwen 2.5 7B} show a steep decline to roughly 73–74\%. At the lower end, models like \textit{Granite-4.0-h-micro} and \textit{Jamba-mini-1.7} struggle with integrated reasoning (below 70\%), and minimal-capacity models such as \textit{Qwen3-30B-A3B} and \textit{Nemotron Nano 9B V2} fail almost entirely. Overall, while large-scale LLMs now excel in comparative system evaluation, achieving coherent integration across diverse UAV mission domains remains a critical next step toward fully autonomous, context-aware reasoning systems.

\subsection{Aggregate Performance and Cross-Style Balance}
\label{subsec:aggregate_results}

Figure~\ref{fig:uavbench_triptych_bss} presents a comparative summary of the top fifteen large language models (LLMs) evaluated under the \texttt{UAVBench\_MCQ} framework. The three subplots respectively illustrate mean accuracy, cross-style consistency (standard deviation of per-style accuracies), and the proposed Balanced Style Score (BSS). Each bar is annotated with the model name and its corresponding performance value, providing a clear view of how overall accuracy and reasoning balance vary across models. The results are ordered by BSS, which rewards models that are both accurate and consistent across reasoning styles.

In panel~(a), the mean accuracy results show that the highest performing models, such as \textit{Qwen3 235B A22B}, \textit{ChatGPT~4o}, and \textit{GPT--5 Chat}, achieve overall accuracies between 80\% and 84\%. This demonstrates that current frontier models maintain strong reasoning capabilities across most UAV mission contexts. Mid-tier systems, including \textit{Qwen3 Max}, \textit{GPT--4.1}, and \textit{Kimi K2}, remain competitive with mean accuracies around 78\%--80\%, whereas smaller or lightweight models tend to cluster near 70\%, confirming the dependence of complex UAV reasoning on model scale and specialization.

Panel~(b) highlights cross-style consistency, expressed as the standard deviation of accuracies across the ten reasoning styles. Lower values indicate more balanced reasoning behavior. Here, the leading models exhibit deviation values below 12\%, signifying stable performance across domains such as physics, planning, ethics, and system integration. In contrast, several mid-range models achieve similar mean accuracies but display higher deviations, implying over-specialization in specific reasoning categories and reduced robustness when generalizing across mission types.

Panel~(c) integrates these dimensions through the Balanced Style Score (BSS), a composite metric that multiplies geometric mean performance by a variance penalty. The results show that \textit{Qwen3 235B A22B} attains the highest BSS of 0.74, followed by \textit{ChatGPT~4o}, \textit{GPT--5 Chat}, and \textit{Qwen3 Max}, each scoring around 0.68. These findings suggest that models combining high accuracy with low cross-style variance achieve the most reliable overall reasoning behavior. Conversely, some models with respectable accuracy but larger variance suffer lower BSS values, reflecting uneven cognitive performance across domains.

Overall, the triptych visualization emphasizes that balanced reasoning, rather than raw accuracy alone, is crucial for evaluating UAV-oriented cognitive competence. High BSS values correspond to models that not only perform well on average but also maintain consistency across all reasoning categories, a property essential for dependable and safe autonomous decision-making.

\section{Conclusion}
\label{sec:conclusion}

This work introduced UAVBench, a large-scale, open benchmark for evaluating autonomous and agentic AI models in UAV systems. \texttt{UAVBench} integrates \textit{50{,}000 validated UAV flight scenarios} constructed through LLM-driven prompt engineering and multi-stage validation, offering a unified schema that encodes environmental, operational, and safety dimensions of UAV missions. On top of this foundation, we developed UAVBench\_MCQ, a structured reasoning benchmark containing \textit{50{,}000 multiple-choice questions} distributed across ten reasoning styles, enabling interpretable and programmatically gradable evaluation of UAV-specific cognition.

Comprehensive evaluation of thirty-two leading LLMs demonstrated that frontier models achieve near-human performance in perception, policy, and physical reasoning, yet remain challenged by multi-agent coordination, energy management, and ethical trade-offs. These findings underscore both the progress and the limitations of current LLMs when applied to safety-critical aerial autonomy. Future extensions of \texttt{UAVBench} will incorporate multimodal sensor data, dynamic simulation rollouts, and temporal reasoning tasks, advancing toward a holistic evaluation framework for embodied, trustworthy, and context-aware UAV intelligence.

\bibliographystyle{IEEEtran}
\bibliography{bibliography} 

@article{wang2024towards,
  title={Towards realistic uav vision-language navigation: Platform, benchmark, and methodology},
  author={Wang, Xiangyu and Yang, Donglin and Wang, Ziqin and Kwan, Hohin and Chen, Jinyu and Wu, Wenjun and Li, Hongsheng and Liao, Yue and Liu, Si},
  journal={arXiv preprint arXiv:2410.07087},
  year={2024}
}

@article{yao2024aeroverse,
  title={Aeroverse: Uav-agent benchmark suite for simulating, pre-training, finetuning, and evaluating aerospace embodied world models},
  author={Yao, Fanglong and Yue, Yuanchang and Liu, Youzhi and Sun, Xian and Fu, Kun},
  journal={arXiv preprint arXiv:2408.15511},
  year={2024}
}

@article{guo2025bedi,
  title={Bedi: A comprehensive benchmark for evaluating embodied agents on uavs},
  author={Guo, Mingning and Wu, Mengwei and He, Jiarun and Li, Shaoxian and Li, Haifeng and Tao, Chao},
  journal={arXiv preprint arXiv:2505.18229},
  year={2025}
}

@article{xiao2025uav,
  title={UAV-ON: A Benchmark for Open-World Object Goal Navigation with Aerial Agents},
  author={Xiao, Jianqiang and Sun, Yuexuan and Shao, Yixin and Gan, Boxi and Liu, Rongqiang and Wu, Yanjing and Gua, Weili and Deng, Xiang},
  journal={arXiv preprint arXiv:2508.00288},
  year={2025}
}

@article{zhao2025urbanvideo,
  title={Urbanvideo-bench: Benchmarking vision-language models on embodied intelligence with video data in urban spaces},
  author={Zhao, Baining and Fang, Jianjie and Dai, Zichao and Wang, Ziyou and Zha, Jirong and Zhang, Weichen and Gao, Chen and Wang, Yue and Cui, Jinqiang and Chen, Xinlei and others},
  journal={arXiv preprint arXiv:2503.06157},
  year={2025}
}

@article{wei2025laura,
  title={LAURA: LLM-Assisted UAV Routing for AoI Minimization},
  author={Wei, Bisheng and Zhang, Ruichen and Jiang, Ruihong and Peng, Mugen and Niyato, Dusit},
  journal={arXiv preprint arXiv:2503.23132},
  year={2025}
}

@inproceedings{sezgin2025llm,
  title={LLM-Powered UAVs: A RAG-Based Approach for Safety-Critical Operations},
  author={Sezgin, An{\i}l and Boyac{\i}, Aytu{\u{g}}},
  booktitle={International Conference on Intelligent and Fuzzy Systems},
  pages={577--584},
  year={2025},
  organization={Springer}
}

@article{zheng2025llm,
  title={LLM Meets the Sky: Heuristic Multi-Agent Reinforcement Learning for Secure Heterogeneous UAV Networks},
  author={Zheng, Lijie and He, Ji and Chang, Shih Yu and Shen, Yulong and Niyato, Dusit},
  journal={arXiv preprint arXiv:2507.17188},
  year={2025}
}

@article{emami2025llm,
  title={LLM-Enabled In-Context Learning for Data Collection Scheduling in UAV-assisted Sensor Networks},
  author={Emami, Yousef and Zhou, Hao and Nabavirazani, SeyedSina and Almeida, Luis},
  journal={arXiv preprint arXiv:2504.14556},
  year={2025}
}

@article{yuan2025next,
  title={Next-Generation LLM for UAV: From Natural Language to Autonomous Flight},
  author={Yuan, Liangqi and Deng, Chuhao and Han, Dong-Jun and Hwang, Inseok and Brunswicker, Sabine and Brinton, Christopher G},
  journal={arXiv preprint arXiv:2510.21739},
  year={2025}
}

@article{li2025efficient,
  title={Efficient Onboard Vision-Language Inference in UAV-Enabled Low-Altitude Economy Networks via LLM-Enhanced Optimization},
  author={Li, Yang and Zhang, Ruichen and Liu, Yinqiu and Liu, Guangyuan and Niyato, Dusit and Jamalipour, Abbas and Wang, Xianbin and Kim, Dong In},
  journal={arXiv preprint arXiv:2510.10028},
  year={2025}
}

@article{wang2025rally,
  title={RALLY: Role-Adaptive LLM-Driven Yoked Navigation for Agentic UAV Swarms},
  author={Wang, Ziyao and Li, Rongpeng and Li, Sizhao and Xiang, Yuming and Wang, Haiping and Zhao, Zhifeng and Zhang, Honggang},
  journal={arXiv preprint arXiv:2507.01378},
  year={2025}
}

@article{zheng2025uav,
  title={UAV Individual Identification via Distilled RF Fingerprints-Based LLM in ISAC Networks},
  author={Zheng, Haolin and Gao, Ning and Cai, Donghong and Jin, Shi and Matthaiou, Michail},
  journal={IEEE Wireless Communications Letters},
  year={2025},
  publisher={IEEE}
}

@inproceedings{wang2025multi,
  title={Multi-uav placement for integrated access and backhauling using llm-driven optimization},
  author={Wang, Yuhui and Farooq, Junaid and Ghazzai, Hakim and Setti, Gianluca},
  booktitle={2025 IEEE Wireless Communications and Networking Conference (WCNC)},
  pages={1--6},
  year={2025},
  organization={IEEE}
}

@article{sekaran2025urbaning,
  title={UrbanIng-V2X: A Large-Scale Multi-Vehicle, Multi-Infrastructure Dataset Across Multiple Intersections for Cooperative Perception},
  author={Sekaran, Karthikeyan Chandra and Geisler, Markus and R{\"o}{\ss}le, Dominik and Mohan, Adithya and Cremers, Daniel and Utschick, Wolfgang and Botsch, Michael and Huber, Werner and Sch{\"o}n, Torsten},
  journal={arXiv preprint arXiv:2510.23478},
  year={2025}
}

@article{ferrag2025reasoning,
  title={Reasoning beyond limits: Advances and open problems for llms},
  author={Ferrag, Mohamed Amine and Tihanyi, Norbert and Debbah, Merouane},
  journal={arXiv preprint arXiv:2503.22732},
  year={2025}
}

@article{hamissi2023survey,
  title={A survey on the unmanned aircraft system traffic management},
  author={Hamissi, Asma and Dhraief, Amine},
  journal={ACM Computing Surveys},
  volume={56},
  number={3},
  pages={1--37},
  year={2023},
  publisher={ACM New York, NY}
}

@article{gong2025safe,
  title={Safe and Economical UAV Trajectory Planning in Low-Altitude Airspace: A Hybrid DRL-LLM Approach with Compliance Awareness},
  author={Gong, Yanwei and Fan, Junchao and Zhang, Ruichen and Niyato, Dusit and Yao, Yingying and Chang, Xiaolin},
  journal={arXiv preprint arXiv:2506.08532},
  year={2025}
}

@article{emami2025frsicl,
  title={FRSICL: LLM-Enabled In-Context Learning Flight Resource Allocation for Fresh Data Collection in UAV-Assisted Wildfire Monitoring},
  author={Emami, Yousef and Zhou, Hao and Gaitan, Miguel Gutierrez and Li, Kai and Almeida, Luis},
  journal={arXiv preprint arXiv:2507.10134},
  year={2025}
}

@article{yan2025hierarchical,
  title={Hierarchical and Collaborative LLM-Based Control for Multi-UAV Motion and Communication in Integrated Terrestrial and Non-Terrestrial Networks},
  author={Yan, Zijiang and Zhou, Hao and Pei, Jianhua and Tabassum, Hina},
  journal={arXiv preprint arXiv:2506.06532},
  year={2025}
}

\end{document}